\documentclass[preprint,12pt]{elsarticle}

\usepackage{amssymb}
\usepackage{amsmath}
\newtheorem{theorem}{Theorem}[section]
\newtheorem{remarque}{Remark}[section]
\RequirePackage{algorithm}
\RequirePackage[noend]{algpseudocode}

\begin{document}

\begin{frontmatter}



\title{Deep Learning for Mean Field Games with non-separable Hamiltonians}

\author[inst1]{Mouhcine Assouli\corref{cor1}}

\affiliation[inst1]{organization={College of Computing, UM6P},
            addressline={Lot 660}, 
            city={Ben Guerir},
            postcode={43150}, 
            country={Morocco}}

\author[inst2]{Badr Missaoui}

\cortext[cor1]{Corresponding author.}

\affiliation[inst2]{organization={College of Computing, UM6P},
            addressline={Lot 660}, 
            city={Ben Guerir},
            postcode={43150}, 
            country={Morocco}}
\begin{abstract}
This paper introduces a new method based on Deep Galerkin Methods (DGMs) for solving high-dimensional stochastic Mean Field Games (MFGs). We achieve this by using two neural networks to approximate the unknown solutions of the MFG system and forward-backward conditions. Our method is efficient, even with a small number of iterations, and is capable of handling up to 300 dimensions with a single layer, which makes it faster than other approaches. In contrast, methods based on Generative Adversarial Networks (GANs) cannot solve MFGs with non-separable Hamiltonians. We demonstrate the effectiveness of our approach by applying it to a traffic flow problem, which was previously solved using the Newton iteration method only in the deterministic case. We compare the results of our method to analytical solutions and previous approaches, showing its efficiency. We also prove the convergence of our neural network approximation with a single hidden layer using the universal approximation theorem.
\end{abstract}


\end{frontmatter}


\section{Introduction}
Mean Field Games (MFGs) are a widely studied topic that can model a variety of phenomena, including autonomous vehicles \cite{huang2019game, shiri2019massive}, finance \cite{cardaliaguet2018mean, casgrain2019algorithmic}, economics \cite{achdou2017income, achdou2014partial, gomes2015economic}, industrial engineering \cite{de2019mean, kizilkale2019integral, gomes2018mean}, and data science \cite{han2018mean, guo2019learning}. MFGs are dynamic, symmetric games where the agents are indistinguishable but rational, meaning that their actions can affect the mean of the population. In the optimal case, the MFG system reaches a Nash equilibrium (NE), in which no agent can further improve their objective. MFGs are described by a system of coupled partial differential equations (PDEs) known as equation 
\begin{equation}\label{1}
\left\{
\begin{array}{rrrrr}
-\partial_t \phi-\nu \Delta \phi + H(x,\rho,\nabla \phi) =0, \ in&E,\\
\partial_t\rho-\nu \Delta \rho - \operatorname{div} \left(\rho \nabla_p H(x,\rho,\nabla \phi) \right)=0, \ in&E, \\
\rho(0,x)=\rho_0(x), \ \ \phi(T,x)=g(x,\rho(T,x)), \ in&\Omega ,
\end{array}
\right.
\end{equation}
where, $E= [0,T] \times \Omega,$  $\Omega$ bounded subset of $\mathbb{R}^d$
 and $g$ denotes the  terminal cost. The Hamiltonian H with separable structure is defined as
\begin{equation}
H(x,\rho,p)=inf_v\{-p.v+L_0(x,v)\}-f_0(x,\rho) 
     =H_0(x,p)-f_0(x,\rho), 
\end{equation}
consisting of a forward-time Fokker-Planck equation (FP) and a backward-time Hamilton-Jacobi-Bellman equation (HJB), which describe the evolution of the population density ($\rho$) and the cost value ($\phi$), respectively. The Hamiltonian $H$ has a separable structure and is defined as the infimum of the Lagrangian function $L_0$, which is the Legendre transform of the Hamiltonian, minus the interaction function $f_0$ between the population of agents. The MFG system also includes initial and terminal conditions, with the initial density $\rho(0,x)$ given by $\rho_0(x)$ and the terminal cost $\phi(T,x)$ given by $g(x,\rho(T,x))$. These conditions apply in the domain $\Omega \subset \mathbb{R}^d$. The solution to system (\ref{1}) exists and is unique under the standard assumptions of convexity of H in the second variable and monotonicity of f and g \cite{lin2020apac}, see also refs \cite{lasry2006jeux, lasry2007mean} for more details. For non-separable Hamiltonians, where the Hamiltonian of the MFG depends jointly on $\rho$ and $p$, the existence and uniqueness of the solution for MFGs of congestion type has been investigated by Achdou and Porretta in \cite{article} and Gomes et al. in \cite{gomes2015short}.\\

One of the main challenges of MFGs is the viscosity problem, in addition to the complexity of the PDEs and forward-backward conditions. Several techniques for solving MFGs are restricted to the deterministic case $(\nu = 0)$.  As an example, \cite{huang2019game}, the authors presented a multigrid preconditioned Newton’s finite difference algorithm for MFG. However, it should be noted that this approach is only applicable when the system is deterministic, and it may not be suitable for systems with viscosity $(\nu > 0)$. While numerical methods do exist for solving the system of PDEs (\ref{1}) \cite{achdou2010mean, benamou2017variational, chow2017algorithm, chow2018algorithm}, it is worth noting that they may not be suitable for deterministic systems and they are not always effective due to  computational complexity, especially in high dimensional problems \cite{hammer1962adaptive, doi:10.1126/science.153.3731.34}. Deep learning methods, such as Generative Adversarial Networks (GANs) \cite{lin2020apac, cao2020connecting}, have been used to address this issue by reformulating MFGs as a primal-dual problem \cite{lasry2007mean, cirant2018variational, benamou2017variational}. This approach uses the Hopf formula in density space \cite{chow2019algorithm} to establish a connection between MFGs and GANs. However, this method requires the Hamiltonian $H$ to be separable in $\rho$ and $p$. In cases where the Hamiltonian is non-separable, such as in traffic flow \cite{huang2019game}, it is not possible to reformulate MFGs as a primal-dual problem. Recently, \cite{lauriere2021policy} proposed a policy iteration algorithm for MFGs with non-separable Hamiltonians for $\nu>0$ using the finite difference method.\\

\emph{Contributions} 
In this work, we present a new method based on DGM for solving stochastic MFG with a non-separable Hamiltonian. Inspired by the work \cite{sirignano2018dgm, raissi2017physics, raissi2018deep}, we approximate the unknown solutions of the system (\ref{1}) by two neural networks trained simultaneously to satisfy each equation of the MFGs system and forward-backward conditions. While the GAN-based techniques are limited to problems with separable Hamiltonians, our algorithm, called MFDGM, can solve high-dimensional MFG systems, including those with separable and non-separable Hamiltonians, as well as both deterministic and stochastic cases. Moreover, we prove the convergence of the neural network approximation with a single layer using a fundamental result of the universal approximation theorem. Then, we test the effectiveness of our MFDGM through several numerical experiments, where we compare our results of MFDGM with previous approaches to assess their reliability. At last, our approach is applied to solve the MFG system of traffic flow accounting for the stochastic case.\\

\emph{Contents} The structure of the rest of the paper is as follows: in Section \ref{Methodology}, we introduce the main description of our approach. Section \ref{Convergence} examines the convergence of our neural network approximation with a single hidden layer. In Section \ref{Related Works}, we present a review of prior methods. Section \ref{Numerical Experiments} investigates the numerical performance of our proposed algorithms. We evaluate our method using a simple analytical solution in Section \ref{Analytic Comparison} and compare it to the previous approach in Section \ref{Comparison }. We also apply our method to the traffic flow problem in Section \ref{secLWR}. Finally, we conclude the paper and discuss potential future work in Section \ref{Conclusion}.

\section{Methodology}\label{Methodology}
\begin{figure}
\centering\includegraphics[width=6.cm]{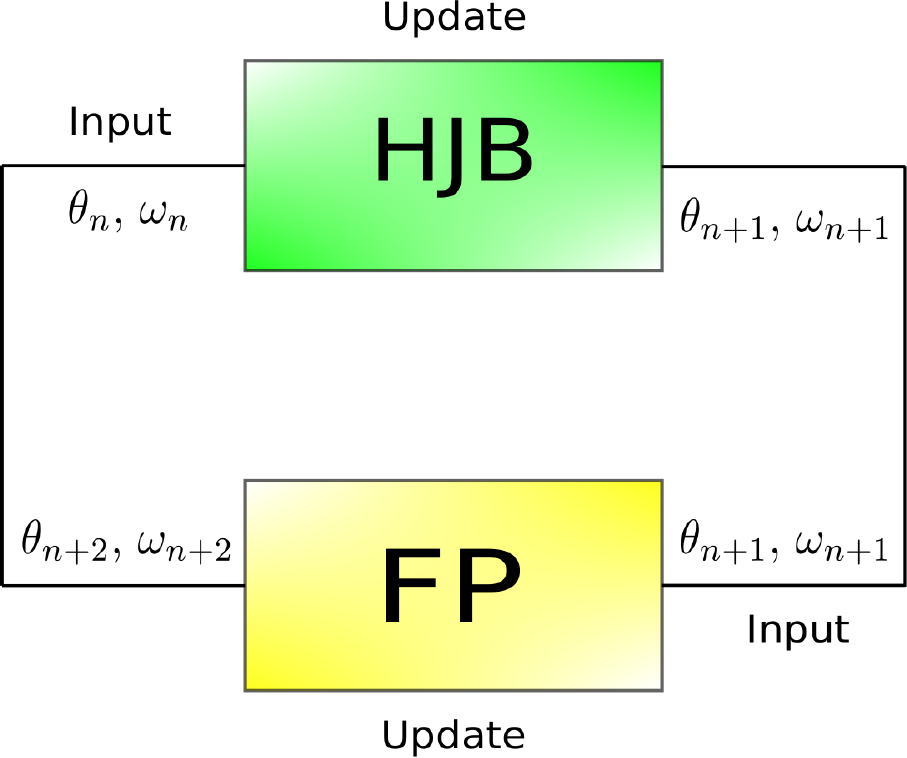}
\caption{The learning mechanism of our method.} 
\label{shema}
\end{figure}

Our method involves using two neural networks, $N_{\theta}$ and $N_{\omega}$, to approximate the unknown variables $\rho$ and $\phi$, respectively. The weights for these networks are $\theta$ and $\omega$. Each iteration of our method involves updating $\rho$ and $\phi$ with the approximations from $N_{\theta}$ and $N_{\omega}$. To optimize the accuracy of these approximations, we use a loss function based on the residual of the first equation (HJB) to update the parameters of the neural networks. We repeat this process using the second equation (FP) and new parameters; see Figure \ref{shema}. Both neural networks are simultaneously trained on the first equation, and the results are then checked in the second equation, where they are fine-tuned until an equilibrium is reached. This equilibrium represents the convergence of the two neural networks and, therefore, the solution to both the Hamilton Jacobi Bellman equations and the Fokker-Planck equation.\\
\\

We have developed a solution for the problem of MFG systems (\ref{1}) that does not rely on the Hamiltonian structure. Our approach involves using a combination of physics-informed deep learning \cite{raissi2017physics} and deep hidden physics models \cite{raissi2018deep} to train our model to solve high-dimensional PDEs that adhere to specified differential operators, initial conditions, and boundary conditions. To train our model, we define a loss function that minimizes the residual of the equation at randomly chosen points in time and space within the domain $\Omega$.\\
We initialize the neural networks as a solution to our system. We let:
\begin{equation}
\phi_{\omega}(t,x)=N_{\omega}(t,x),  \ \ \ \rho_{\theta}(t,x)=N_{\theta}(t,x).
\end{equation}
Our training strategy starts by solving (HJB). We compute the loss (\ref{los}) at randomly sampled points $\{(t_{b},x_{b})\}_{b=1}^{B}$ from $E$, and  $\{x_{s}\}_{s=1}^{S}$ from $\Omega$ according to respective
probability densities $\mu_1$ and $\mu_2$.
\begin{equation}\label{los}
\textsc{Loss}_{total}^{(HJB)}=\textsc{Loss}^{(HJB)}+\textsc{Loss}_{cond}^{(HJB)}, 
\end{equation}
where
\begin{align*}
 \textsc{Loss}^{(HJB)}&=\frac{1}{B} \sum_{b=1}^{B}\Big| \partial_t \phi_{\omega}(t_{b},x_{b})
     + \nu \Delta \phi_{\omega}(t_{b},x_{b})
     \\ &\quad - H(x_{b},\rho_{\theta}(t_{b},x_{b}),\nabla \phi_{\omega}(t_{b},x_{b}))\Big|^2,  
\end{align*}
and
$$\textsc{Loss}_{cond}^{(HJB)}=\frac{1}{S} \sum_{s=1}^{S} \Big|\phi_{\omega}(T,x_{s})-g(x_{s},\rho_{\theta}(T,x_{s}))\Big|^2.$$
We then update the weights of $\phi_{\omega}$ and  $\rho_{\theta}$ by back-propagating the loss (\ref{los}). We do the same to (FP) with the updated weights. We compute (\ref{los2}) at randomly sampled points $\{(t_{b},x_{b})\}_{b=1}^{B}$ from $E$, and  $\{x_{s}\}_{s=1}^{S}$ from $\Omega$ according to respective probability densities $\mu_1$ and $\mu_2$.\\
\begin{equation}\label{los2}
    \textsc{Loss}_{total}^{(FP)}=\textsc{Loss}^{(FP)}+\textsc{Loss}_{cond}^{(FP)},
\end{equation}
where
\begin{align*}
\textsc{Loss}^{(FP)}&=\frac{1}{B} \sum_{b=1}^{B} \Big|\partial_t \rho_{\theta}(t_{b},x_{b})
     - \nu \Delta \rho_{\theta}(t_{b},x_{b})\\
     &\quad- \operatorname{div} \left(\rho_{\theta}(t_{b},x_{b})\nabla_pH(x_{b},\rho_{\theta}(t_{b},x_{b}),\nabla \phi_{\omega}(t_{b},x_{b})) \right)\Big|^2,
\end{align*}
and
\begin{equation*}
   \textsc{Loss}_{cond}^{(FP)}=\frac{1}{S} \sum_{s=1}^{S}\Big|\rho_{\theta}(0,x_{s})-\rho_0(x_{s})\Big|^2. 
\end{equation*}
Finally, we update the weights of $\phi_{\omega}$ and $\rho_{\theta}$ by back-propagating the loss (\ref{los2}); see Algorithm \ref{alg1}.
\begin{algorithm}[hbt!]
\begin{algorithmic}
\Require $H$ Hamiltonian, $\nu $ diffusion parameter, $g$ terminal cost. 
\Require Initialize neural networks $ N_{\omega_0}$ and $N_{\theta_0}$\\
\textbf{Train}
\For{n=0,1,2...,K-2}
     \State Sample batch $\{(t_{b},x_{b})\}_{b=1}^{B}$ from $E$, and  $\{x_{s}\}_{s=1}^{S}$ from $\Omega$
     \State $\textsc{L}^{(HJB)}\leftarrow\frac{1}{B} \sum_{b=1}^{B}\Big| \partial_t \phi_{\omega_n}(t_{b},x_{b})
     + \nu \Delta \phi_{\omega_n}(t_{b},x_{b})$
      \State $\quad \quad \quad \quad \quad- H(x_{b},\rho_{\theta_n}(t_{b},x_{b}),\nabla \phi_{\omega_n}(t_{b},x_{b}))\Big|^2.$
     \State $\textsc{L}_{cond}^{(HJB)}\leftarrow \frac{1}{S} \sum_{s=1}^{S} \Big|\phi_{\omega_n}(T,x_{s})-g(x_{s},\rho_{\theta_n}(T,x_{s}))\Big|^2.$
     \State Backpropagate $\textsc{Loss}_{total}^{(HJB)}$  to $\omega_{n+1}$, $\theta_{n+1}$ weights.
     \State \\
     \State Sample batch $\{(t_{b},x_{b})\}_{b=1}^{B}$ from $E$, and  $\{x_{s}\}_{s=1}^{S}$ from $\Omega$.
     \State $\textsc{L}^{(FP)}\leftarrow \frac{1}{B} \sum_{b=1}^{B} \Big|\partial_t \rho_{\theta_{n+1}}(t_{b},x_{b})
     - \nu \Delta \rho_{\theta_{n+1}}(t_{b},x_{b})$
     \State $\quad \quad \quad \quad- \operatorname{div} (\nabla_pH(x_{b},\rho_{\theta_{n+1}}(t_{b},x_{b}),\nabla \phi_{\omega_{n+1}}(t_{b},x_{b}))$
    \State $ \quad \quad \quad \quad \quad \times\rho_{\theta_{n+1}}(t_{b},x_{b}) )\Big|^2.$ 
     \State $\textsc{L}_{cond(FP)}\leftarrow\frac{1}{S} \sum_{s=1}^{S}\Big|\rho_{\theta{n+1}}(0,x_{s})-\rho_0(x_{s})\Big|^2.$
     \State Backpropagate  $\textsc{Loss}_{total}^{(FP)}$ to $\omega_{n+2}$ $\theta_{n+2}$ weights.
     \State \\
\EndFor 
\Return $\theta_{K}$, $\omega_{K}$    
\end{algorithmic}
\caption{MFDGM}\label{alg1}
\end{algorithm}

\section{Convergence}\label{Convergence}
Following the steps of \cite{sirignano2018dgm}, this section presents theoretical results that guarantee the existence of a single layer feedforward neural networks $\rho_{\theta}$ and $\phi_{\omega}$ which can universally approximate the solutions of (\ref{1}). Denote
\begin{equation}
L_1(\rho_{\theta},\phi_{\omega})=\Big\|\mathcal{H}_1(\rho_{\theta},\phi_{\omega})\Big\|^2_{L^2(E)}
     + \Big\|\phi_{\omega}(T,x)-\phi(T, x)\Big\|^2_{L^2(\Omega)},
\end{equation}
where
\begin{equation*}
\mathcal{H}_1(\rho_{\theta},\phi_{\omega})=\partial_t \phi_{\omega}(t,x)
     + \nu \Delta \phi_{\omega}(t,x)
      - H(x,\rho_{\theta}(x,t),\nabla \phi_{\omega}(t,x)).
\end{equation*}
\begin{equation}
L_2(\rho_{\theta},\phi_{\omega})
=\Big\|\mathcal{H}_2(\rho_{\theta},\phi_{\omega})\Big\|^2_{L^2(E)}
     + \Big\|\rho_{\theta}(0,x)-\rho_0(x)\Big\|^2_{L^2(\Omega)},
\end{equation}
and 
\begin{equation*} \mathcal{H}_2(\rho_{\theta},\phi_{\omega})=\partial_t \rho_{\theta}(t,x)
     - \nu \Delta \rho_{\theta}(t,x)\\ 
     - \operatorname{div} \left(\rho_{\theta}(t,x)\nabla_pH(x,\rho_{\theta}(t,x),\nabla \phi_{\omega}(t,x))\right).
\end{equation*}
Denote $||f(x)||_{L^2(E)}=\left(\int_E|f(x)|^2 d\mu(x)\right)^{\frac{1}{2}}$ the norm on $L^2$ and $\mu$ is a positive probability density on $E$. We recall some key definitions of functional spaces, which  we will explore later on,
$\mathcal{C}^{1,2}(E)$ is the set of all continuous functions in $E$ having continuous derivatives $u_t, u_x, u_{xx}$ in $E.$
$\mathcal{C}^{1/2,l}(E)$ is the set of all continuous functions in $E$ satisfying Holder condition  in  t  with  exponent $1/2$ and $l$ Lipschitz condition in $x.$The aim of our approach is to identify a set of  parameters $\theta$ and $\omega$ such that the functions $\rho_{\theta}(x,t)$ and  $\phi_{\omega}(x,t)$ minimizes the error $L_1(\rho_{\theta},\phi_{\omega})$ and $L_2(\rho_{\theta},\phi_{\omega})$. If $L_1(\rho_{\theta},\phi_{\omega})=0$ and $L_2(\rho_{\theta},\phi_{\omega})=0,$ then $\rho_{\theta}(t,x)$ and  $\phi_{\omega}(t,x)$ are solutions to (\ref{1}). To prove the convergence of the neural networks, we use the results \cite{hornik1991approximation} on the universal approximation of functions and their derivatives. Define the class of neural networks with a single hidden layer and $n$ hidden units,
\begin{equation*}
\mathcal{N}^{n}(\sigma)=\Big\{\Phi(t,x): \mathbb{R}^{1+d} \mapsto \mathbb{R}:\\ \Phi(t, x)
    =\sum_{i=1}^{n} \beta_{i} \sigma\left(\alpha_{1, i} t+\sum_{j=1}^{d} \alpha_{j, i} x_{j}+c_{j}\right)\Big\},
\end{equation*}
where $\sigma$ is the common activation function of the
hidden units and,   $$\theta=\left(\beta_{1}, \cdots, \beta_{n}, \alpha_{1,1}, \cdots, \alpha_{d, n}, c_{1}, c_{1}, \cdots, c_{n}\right) \in \mathbb{R}^{2 n+n(1+d)},$$
the vector of the parameter to be learned. The set of all functions implemented by such a network with a single hidden layer and $n$ hidden units is
\begin{equation}
   \mathcal{N}(\sigma)=\bigcup_{n\geq 1}\mathcal{N}^n(\sigma),
\end{equation}
We consider $\mathbb{E}$ a compact subset of $\mathbb{R}^{d+1}$, from \cite[Th~3]{hornik1991approximation}. We know that if $\sigma \in \mathcal{C}^{2}\left(\mathbb{R}^{d+1}\right)$ is non constant and bounded, then $\mathcal{N}(\sigma)$ is uniformly 2-dense on $E$. This means by \cite[Th~2]{hornik1991approximation} that for all $u \in \mathcal{C}^{1,2}\left([0, T] \times \mathbb{R}^{d}\right)$ and $\epsilon>0$, there is $f_{\theta} \in \mathcal{N}(\sigma)$ such that:
\begin{equation}
\sup _{(t, x) \in \mathbb{E}}\left|\partial_{t} u(t, x)-\partial_{t} f_{\theta}(t, x)\right|
       +\max _{|a| \leq 2} \sup _{(t, x) \in \mathbb{E}}\left|\partial_{x}^{(a)} u(t, x)-\partial_{x}^{(a)} f_{\theta}(t, x )\right|<\epsilon. 
\end{equation}
To prove the convergence of our algorithm, we make the following assumptions,
\begin{itemize}
    \item {\bf(H1):} $E$  compact and consider the measures $\mu_{1} \ \text{and} \ \mu_{2}$ whose support is contained in $E \ \text{and} \ \Omega$ respectively.
    \item {\bf(H2):} System (\ref{1}) has a unique solution $(\phi, \rho) \in \mathcal{X} \times \mathcal{X}$ such that:
\begin{equation*}
    \begin{split}
      \mathcal{X}&=\Big\{u(t,x) \in \mathcal{C}\left(\bar{E}\right) \bigcap \mathcal{C}^{1+\eta/2,2+\eta}\left(E\right)\\ &\quad\quad\quad\text{with} \ \ \eta \in(0,1)
      \text{and that} \sup _{(t, x) \in E} \sum_{k=1}^{2}\left|\nabla_{x}^{(k)} u(t, x)\right|<\infty \Big\}.
    \end{split}
    \end{equation*}
    \item {\bf(H3):}  $H,  \ \nabla_pH, \ \nabla_{pp}H, \ \nabla_{\rho p}H$ are locally
Lipschitz continuous in $(\rho, p)$  with Lipschitz constant
that can have at most polynomial growth in $\rho$ and $p$, uniformly with respect to $t, x.$ 
\end{itemize}
\begin{remarque}\label{rmk}
It is important to note that the nonlinear term of $L_2$ can be simplified as follows,
\begin{equation*}
    \begin{split}
    \operatorname{div}(\rho \nabla_pH(x,\rho,\nabla\phi))&=\nabla_pH(x,\rho,\nabla\phi)\nabla\rho  +\nabla_{p\rho}H(x,\rho,\nabla\phi)\nabla\rho.\rho\\
    &\quad+ \sum_{i,j}\nabla_{p_ip_j}H(x,\rho,\nabla\phi)(\partial_{x_jx_i}\phi)\rho.
    \end{split}
\end{equation*}
\end{remarque}
\begin{theorem}\label{the}
Let consider $\mathcal{N}(\sigma)$ where $\sigma$ is $ \mathcal{C}^{2}\left(\mathbb{R}^{d+1}\right)$,  non constant and bounded. Suppose {\bf (H1)}, {\bf (H2)}, {\bf (H3)} hold. Then for every $\epsilon > 0$, there exists two positives constant $C_1, C_2 >0$ and there exists two functions $(\rho_{\theta}, \phi_{\omega}) \in \mathcal{N}(\sigma) \times \mathcal{N}(\sigma)$, such that, $$L_i(\rho_{\theta}, \phi_{\omega})\leq C_i \epsilon, \ \ \  \ \ \ \ \text{for} \ \ \ i=\{1, 2\}.$$ 
\end{theorem}

The proof of this theorem is in \ref{Theorem 3.1.}.\\
\\
Now we have $L_1(\rho_{\theta}^n, \phi_{\omega}^n)\mapsto 0$, and $L_2(\rho_{\theta}^n, \phi_{\omega}^n)\mapsto 0$ as $n \mapsto \infty$ but it does not necessarily imply that
$(\rho_{\theta}^n, \phi_{\omega}^n)\mapsto(\rho, \omega)$ is the unique solution.\\
We now prove, under stronger conditions, the convergence of the neural network, $\left(\rho_{\theta}^{n}, \phi_{w}^{n}\right)$ to the solution $(\rho, \phi)$ of the system (\ref{1}) as $n \rightarrow \infty$. It is important to add boundary conditions to prove this convergence  that will help us to ensure the problem is well-imposed, and to benefit from the results of the \cite{sirignano2018dgm, boccardo1997nonlinear}. 
To avoid some difficulties, we add homogeneous boundary conditions that assume the solution is vanishing on the boundary. We assume $\Omega$ a bounded, open subset of $\mathbb{R}^d$, the MFG system (\ref{1}) writes
\begin{equation}\label{11}
\left\{
\begin{array}{rrrrr}
-\partial_t \phi-\nu \operatorname{div}\left(a_{1}( \nabla \phi)\right)+\gamma(\rho, \nabla \phi)=0, \ in&\Omega_T,\\
\partial_t\rho-\nu \operatorname{div}\left(a_{2}(\nabla \rho)\right) - \operatorname{div}\left(a_{3}( \rho, \nabla \phi)\right)=0, \ in&\Omega_T, \\
\rho(0,x)=\rho_0(x), \ \ \phi(T,x)=g(x,\rho(T,x)), \ in&\Omega, \\
\rho(t, x)=\phi(t, x)=0, \ in&\Gamma ,
\end{array}
\right.
\end{equation}
Where,  $\Omega_T=(0,T)\times \Omega$, $\Gamma=(0,T)\times \partial\Omega$ and 
\begin{equation*}
\begin{array}{l}
a_{1}(t,x,\nabla \phi)=\nabla \phi, \\
a_{2}(t,x,\nabla \rho)=\nabla \rho, \\
a_{3}(t,x,\rho, \nabla \phi)=\rho \nabla p H(x,\rho, \nabla \phi), \\
\gamma(t,x,\rho, \nabla \phi)=H(x, \rho, \nabla \phi),
\end{array}
\end{equation*}
\\
$a_{1}: \Omega_T  \times \mathbb{R}^{d} \rightarrow \mathbb{R}^{d}$, $a_{2}: \Omega_T  \times \mathbb{R}^{d} \times \mathbb{R}^{d} \rightarrow \mathbb{R}^{d}$, $a_{3}: \Omega_T \times \mathbb{R} \times \mathbb{R}^{d} \rightarrow \mathbb{R}^{d}$ and $\gamma: \Omega_T \times \mathbb{R} \times \mathbb{R}^{d} \rightarrow \mathbb{R}$ are Caratheodory functions.\\ 
Then we introduce the approximate problem of the system (\ref{11}) as
\\
\begin{equation}\label{12}
\left\{
\begin{array}{rrrrr}
-\partial_t \phi_{\omega}^{n}-\nu \operatorname{div}\left(a_{1}( \nabla \phi_{\omega}^{n})\right)+\gamma(\rho_{\theta}^{n}, \nabla \phi_{\omega}^{n})=0, \ in&\Omega_T,\\
\partial_t\rho_{\theta}^{n}-\nu \operatorname{div}\left(a_{2}( \nabla \rho_{\theta}^{n})\right) - \operatorname{div}\left(a_{3}(\rho_{\theta}^{n}, \nabla \phi_{\omega}^{n}\right)=0, \ in&\Omega_T, \\
\rho_{\theta}^{n}(0,x)=\rho_0(x), \ \ \phi_{\omega}^{n}(T,x)=g(x,\rho_{\theta}^{n}(T,x)), \ in&\Omega, \\
\rho_{\theta}^{n}(t, x)=\phi_{\omega}^{n}(t, x)=0. \ in&\Gamma ,
\end{array}
\right.
\end{equation}
\\
Let us first introduce some definitions.\\
Let $r \geq 1$. In the sequel we denote by $L^{r}\left(0, T ; W_{0}^{1, r}(\Omega)\right)$ the set of functions $u$ such that $u \in L^{r}\left(\Omega_T\right)$, $u(t,\cdot) \in W_{0}^{1, r}(\Omega)$. The space $L^{r}\left(0, T ; W_{0}^{1, r}(\Omega)\right)$ is equipped with the norm
$$
\|u\|_{L^{r}\left(0, T ; W_{0}^{1, r}(\Omega)\right)}:=\left(\int_{0}^{T} \int_{\Omega}|\nabla u(x, t)|^{r} d x d t\right)^{\frac{1}{r}},
$$
is a Banach space.
For $s, r \geq 1$, the space $V_{0}^{s, r}\left(\Omega_T\right):=L^{\infty}\left(0, T ; L^{s}(\Omega)\right) \cap L^{r}\left(0, T ; W_{0}^{1, r}(\Omega)\right)$ endowed with the norm
$$
\|\varphi\|_{V_{0}^{s, r}\left(\Omega_T\right)}:=\operatorname{ess} \sup _{0 \leq t \leq T}\|\varphi(., t)\|_{L^{s}(\Omega)}+\|\varphi\|_{L^{r}\left(0, T ; W_{0}^{1, r}(\Omega)\right)},
$$
is also a Banach space.\\
To ensure this convergence, we impose the following set of assumptions on $\gamma$ and, by similarity, on $a_3$ to leverage the results of Paper \cite{sirignano2018dgm, boccardo1997nonlinear}.
\begin{itemize}
    \item {\bf (H4):} There is a constant $\mu>0$ and positive functions $\kappa(t, x), \lambda(t, x)$ such that for all $(t, x) \in$ $\Omega_T$, we have
$$
\|\gamma(t, x, \rho, p)\| \leq \mu(\kappa(t, x)+\|p\|), \text { and }|\gamma(t, x, \rho, p)| \leq \lambda(t, x)\|p\|,
$$
with $\kappa \in L^{2}\left(\Omega_T\right), \lambda \in L^{d+2}\left(\Omega_T\right).$
\item {\bf (H5):} $\gamma(t, x, \rho, p)$ are Lipschitz continuous in $(t, x, \rho, p) \in \Omega_T \times \mathbb{R} \times \mathbb{R}^{d}$ uniformly on compacts of the form $\left\{(t, x) \in \Omega_T,|\rho| \leq C,|p| \leq C\right\}$.
\item {\bf (H6):} There is a positive constant $\alpha>0$ such that
$$
\gamma(t, x, \rho, p) p \geq \alpha|p|^{2}, 
$$
and
$$
\left\langle\gamma\left(t, x, u, p_1\right)-\gamma\left(t, x, u, p_2\right), p_1-p_2\right\rangle>0, \text { for every } p_1, p_2 \in \mathbb{R}^d, p_1 \neq p_2 .
$$

\end{itemize}
\begin{theorem}
Under previous assumptions {\bf (H4)-(H6)}, if we assume that (\ref{11}) has a unique bounded solution $(\phi, \rho )\in V_{0}^{2, 2} \times V_{0}^{2, 2}$, then $(\phi_{\omega}^n, \rho_{\theta}^n)$  converge to $(\phi, \rho )$ strongly in $L^{p}\left(\Omega_T\right) \times L^{p}\left(\Omega_T\right)$ for every $p<2$.
\end{theorem}

The proof of this theorem is in \ref{Theorem 3.2.}.

\section{Related Works}\label{Related Works}

{\bf GANs:} Generative adversarial networks, or GANs, are a class of machine learning models introduced in 2014 \cite{goodfellow2020generative} that have been successful in generating images and processing data \cite{denton2015deep, reed2016generative, radford2015unsupervised}. In recent years, there has been increasing interest in using GANs for financial modeling as well \cite{wiese2019deep}. GANs consist of two neural networks, a generator network, and a discriminator network, that work against each other in order to generate samples from a specific distribution. As described in various sources \cite{goodfellow2020generative, dukler2019wasserstein, villani2021topics}, the goal is to reach equilibrium for the following problem,
\begin{equation}\label{gan6}
\min\limits_{G}\max\limits_{D}\Big\{\mathbb{E}_{x \sim P_{data}(x)}[log(D(x)]\\
     + \mathbb{E}_{z \sim P_{g}(z)}[log(1-D(G(z))]\Big\},
\end{equation}
where $P_{data}(x)$ is the original data and $P_{g}(z)$ is the noise data. In (\ref{gan6}), the goal is to minimize the generator's output (G) and maximize the discriminator's output (D). This is achieved by comparing the probability of the original data $P_{data}(x)$ being correctly identified by the discriminator D with the probability of the generated data G produced by the generator using noise data $P_{g}(z)$ being incorrectly identified as real by the discriminator $1-D(G(z))$. Essentially, the discriminator is trying to accurately distinguish between real and fake data, while the generator is attempting to create fake data that can deceive the discriminator.\\

{\bf   APAC-Net:} In \cite{lin2020apac}, the authors present a method (APAC-Net) based on GANs for
solving high-dimensional MFGs in the stochastic case. They use of the Hopf formula in density space to reformulate the MFGs as a  saddle-point problem  given by,
\\
\begin{equation}\label{eqGM4}
\begin{split}
\inf\limits_{\rho(x,t)}\sup\limits_{\phi(x,t)}&\Big\{\mathbb{E}_{z\sim P(z),t \sim Unif[0,T]} [\partial_t \phi(\rho(t,z),t)+ \nu \Delta \phi(\rho(t,z),t
      )\\  &\quad- H(\rho(t,z),\nabla \phi)]
      + \mathbb{E}_{z\sim P(z)} \phi(0,\rho(0,z)) - \mathbb{E}_{x \sim \rho_T} \phi(T,x) \Big \}, 
\end{split}
\end{equation}
where 
$$H(x,p)=inf_v\{-p.v+L(x,v)\}.$$
In this case, we have a connection between the GANs and MFGs, since (\ref{eqGM4}) allows them to reach the Kantorovich-Rubenstein dual formulation of Wasserstein GANs \cite{villani2021topics} given by,
\\
\begin{equation}\label{gan6bis}
\begin{array}{cc}
     \min\limits_{G}\max\limits_{D} \{\mathbb{E}_{x \sim P_{data}(x)}[(D(x)] - \mathbb{E}_{z \sim P_{g}(z)}[(D(G(z))]\},  \\
     s.t. \ \ ||\nabla D||\leq 1. 
\end{array}
\end{equation}
\\
Finally, we can use an algorithm similar to GANs to solve the problems of MFGs. Unfortunately, we
notice that the Hamiltonian in this situation has a separable structure. Due to this, we cannot solve the  MFG-LWR system (to be detailed in section \ref{secLWR}). In general, we cannot solve the MFGs problems, where its Hamiltonian is non-separable, since we cannot reformulate MFGs as \ref{eqGM4}.
\\

{\bf MFGANs:} In \cite{cao2020connecting, lin2020apac}, the connection between GANs and MFGs is demonstrated by the fact that equation (\ref{eqGM4}) allows them to both reach the Kantorovich-Rubinstein dual formulation of Wasserstein GANs, as described in reference \cite{villani2021topics}. This is shown in equation (\ref{gan6}), which can be solved using an algorithm similar to those used for GANs. However, it is not possible to solve MFGs problems with non-separable Hamiltonians, as they cannot be reformulated as in equation (\ref{eqGM4}). This is because the Hamiltonian in these cases has a separable structure, which prevents the solution of the MFG-LWR system (to be discussed in section \ref{secLWR}).
\\

{\bf DGM-MFG:} In \cite{carmona2021deep}, section 4 discusses the adaptation of the DGM algorithm to solve mean field games, referred to as DGM-MFG. This method is highly versatile and can effectively solve a wide range of partial differential equations due to its lack of reliance on the specific structure of the problem. Our own work is similar to DGM-MFG in that we also utilize neural networks to approximate unknown functions and adjust parameters to minimize a loss function based on the PDE residual, as seen in \cite{carmona2021deep} and \cite{cao2020connecting}. However, our approach, referred to as MFDGM, differs in the way it is trained. Instead of using the sum of PDE residuals as the loss function and SGD for optimization, we define a separate loss function for each equation and use ADAM for training, following the approach in \cite{cao2020connecting}. This modification allows for faster and more accurate convergence. 
\\

{\bf Policy iteration Method:} To the best of our knowledge, \cite{lauriere2021policy} was the first to successfully solve systems of mean field game partial differential equations with non-separable Hamiltonians. They proposed two algorithms based on policy iteration, which involve iteratively updating the population distribution, value function, and control. These algorithms only require the solution of two decoupled, linear PDEs at each iteration due to the fixed control. This approach reduces the complexity of the equations, but it is limited to low-dimensional problems due to the computationally intensive nature of the method. In contrast, our method utilizes neural networks to solve the HJB and FP equations at each iteration, allowing for updates to the population distribution and value function in each equation without the limitations of \cite{lauriere2021policy}.

\section{Numerical Experiments}\label{Numerical Experiments}
To evaluate the effectiveness of the proposed algorithm [\ref{alg1}], we use the example provided in \cite{lin2020apac}, as it has an explicitly defined solution structure that allows for easy numerical comparison. We compare the performance of MFDGM, APAC-Net's MFGAN, and DGM-MFG on the same data to assess their reliability. Additionally, we apply MFDGM to the traffic flow problem [\ref{5}], which is characterized by its non-separable Hamiltonian [\ref{nonsep}], to determine its ability to solve this type of problem.
\subsection{Analytic Comparison}\label{Analytic Comparison}
We test our method by comparing it to a simple example of the analytic solution used to test the effectiveness of Apac-Net  \cite{lin2020apac}. For the sake of simplicity,  we take  the spatial domain $\Omega=[-2,2]^d$,  the final time $T=1$, and without congestion $(\gamma=0)$. For
\begin{equation}
\begin{array}{cc}
     H_0(x,p)=\frac{||p||^2}{2}-\beta \frac{||x||^2}{2}, \ \ \ f_0(x,\rho)=\gamma \texttt{ln}(\rho),  \\
     g(x)=\alpha\frac{||x||^2}{2}-(\nu d \alpha +\gamma\frac{d}{2}ln \frac{\alpha}{2\pi\nu}),
\end{array}
\end{equation}
and $\nu=\beta=1$,
where $$\alpha=\frac{-\gamma +\sqrt{\gamma^2+4\nu^2 \beta}}{2\nu}=1.$$
The corresponding MFG system is:
\\
\begin{equation}\label{test}
\left\{
\begin{array}{rrrrr}
-\partial_t \phi- \Delta \phi + \frac{||\nabla \phi||^2}{2}- \frac{||x||^2}{2} =0, \\
\partial_t\rho- \Delta \rho - \operatorname{div} \left(\rho \nabla \phi \right)=0,  \\
\rho(0,x)=(\frac{1}{2\pi})^{\frac{d}{2}}e^{-\frac{ ||x||^2}{2}}, \\ \phi(T,x)=\frac{x^2}{2}-d, 
\end{array}
\right.
\vspace{-0.25cm}
\end{equation}
and the explicit formula is given by 
\begin{equation}
\begin{array}{cc}
     \phi(t,x)=\frac{||x||^2}{2}-d.t, \\
     \rho(t,x)=(\frac{1}{2\pi})^{\frac{d}{2}}e^{-\frac{ ||x||^2}{2}}.
\end{array}
\end{equation}

\begin{figure}
\centering\includegraphics[width=5.cm]{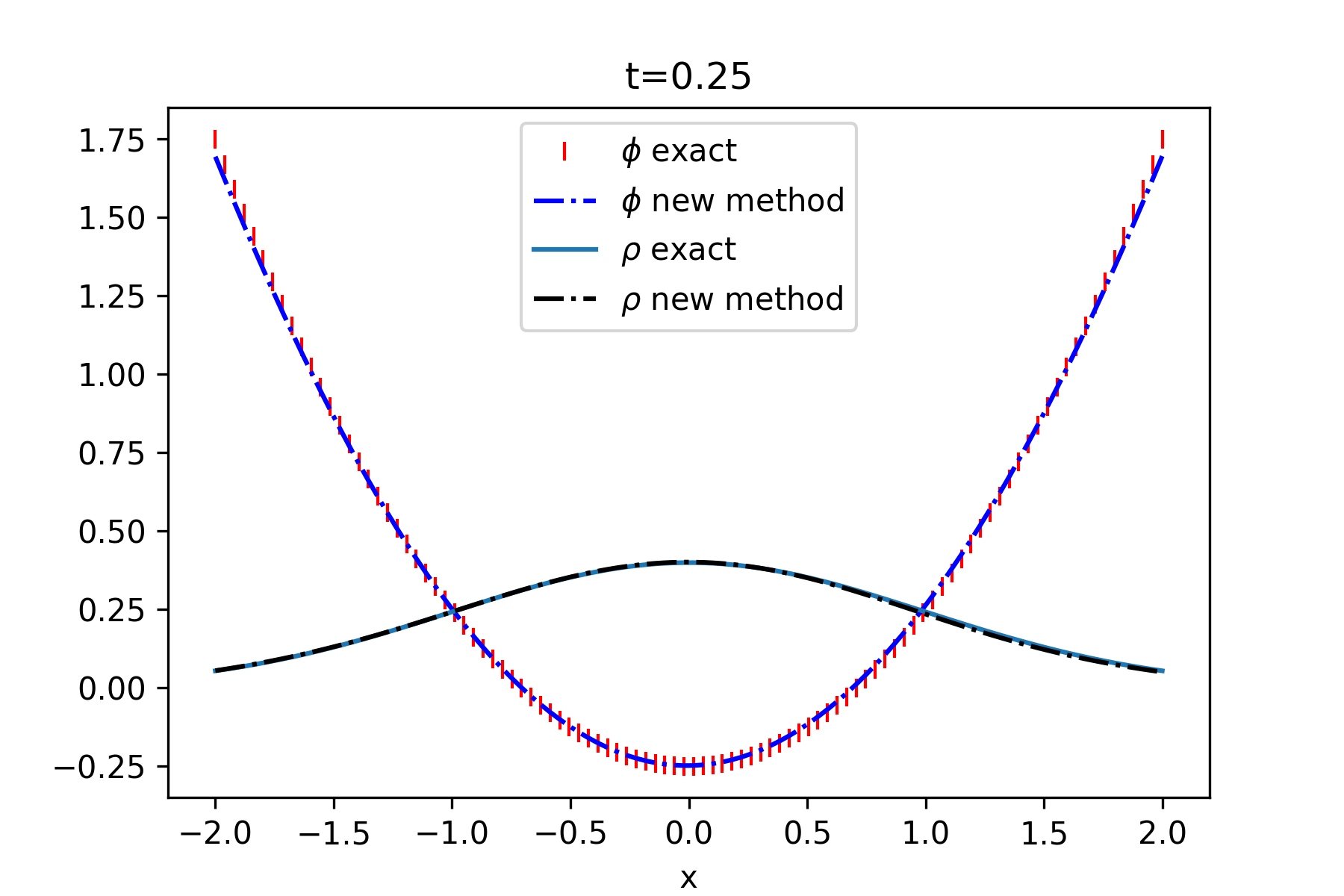}
\centering\includegraphics[width=5.cm]{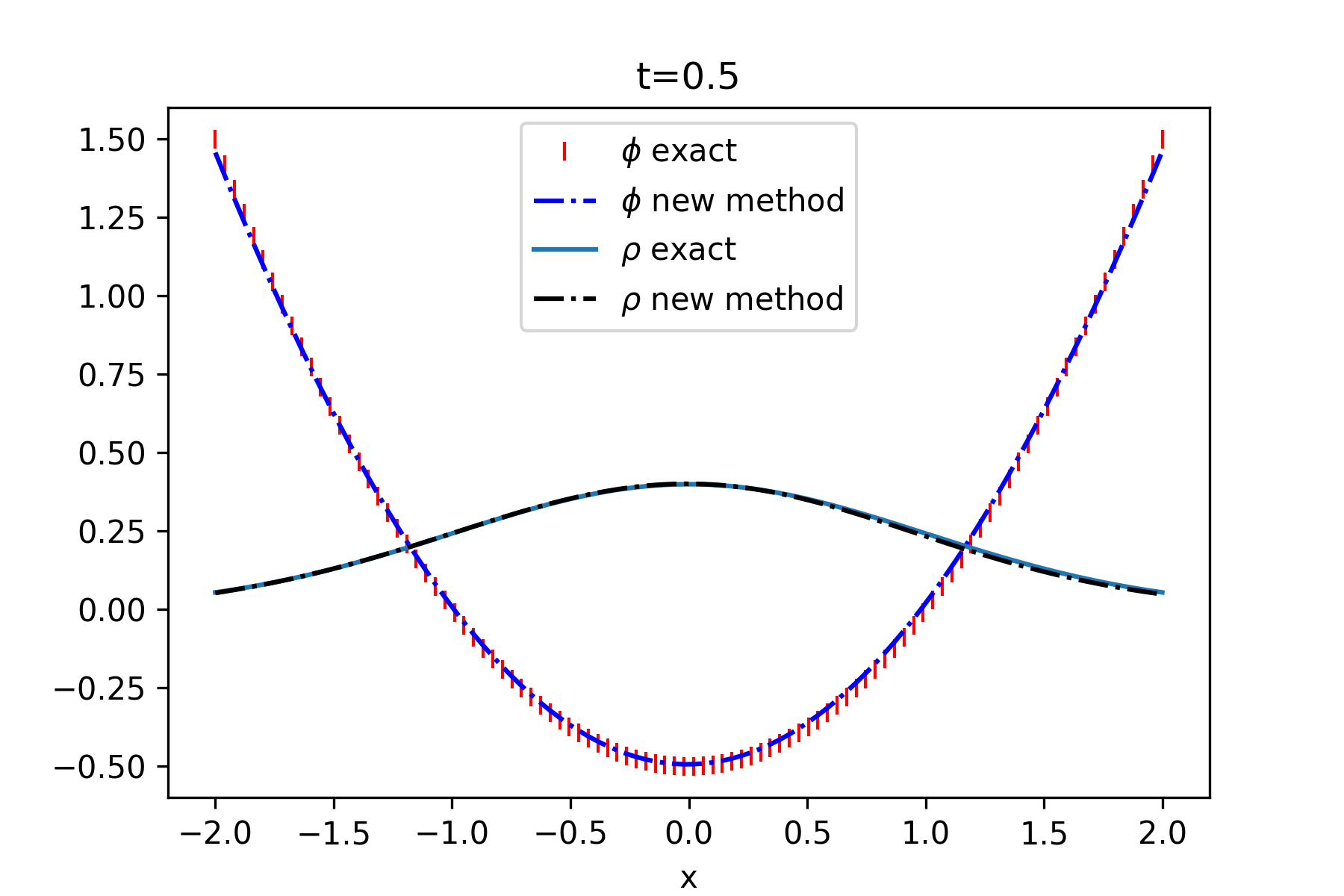}
\centering\includegraphics[width=5.cm]{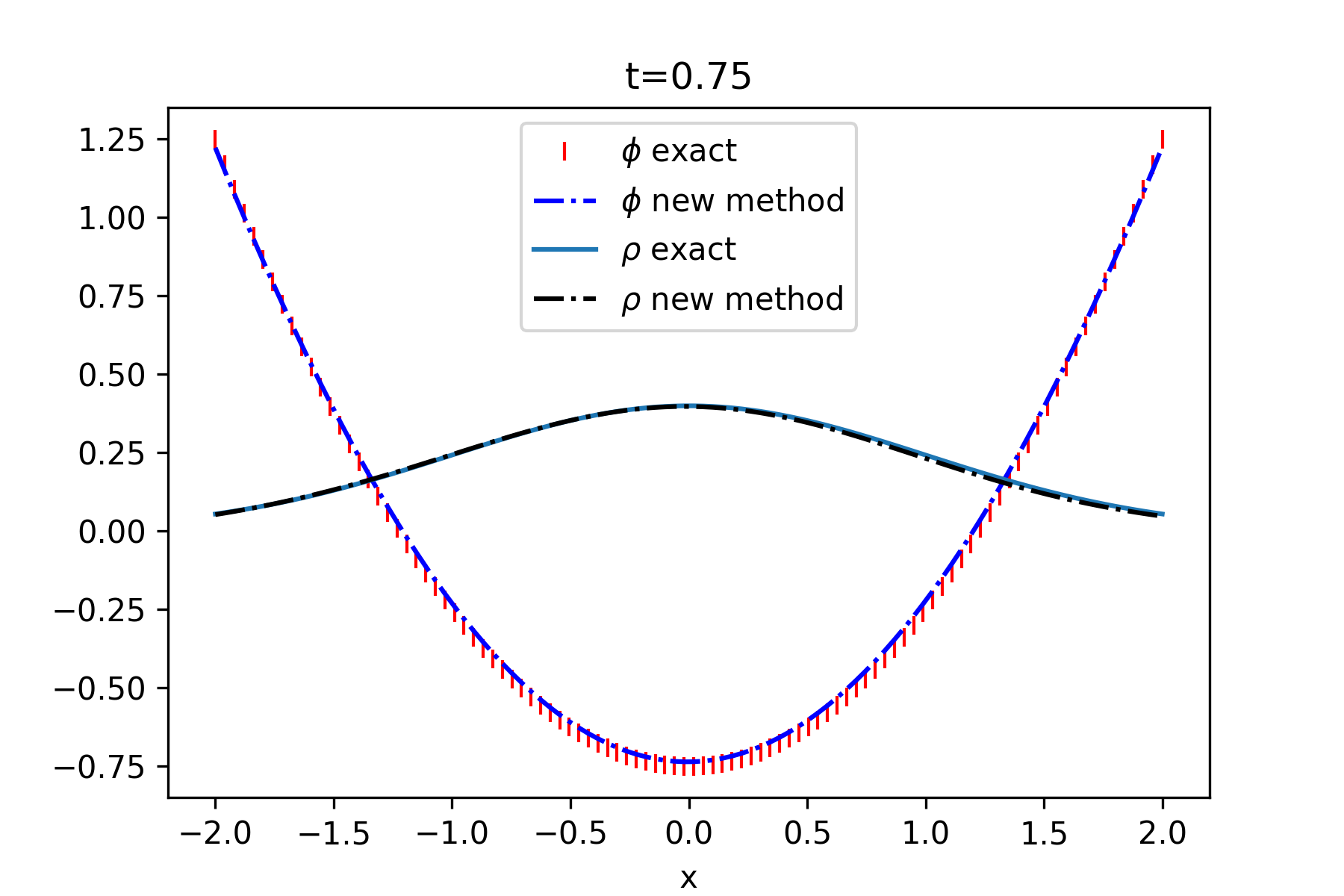}
\caption{The exact solution and prediction calculated by MFDGM in dimension one at t=(0.25, 0.5, 0.75 ).} 
\label{fig1}
\end{figure}

\begin{figure}[htbp]
\centering\includegraphics[width=6.cm]{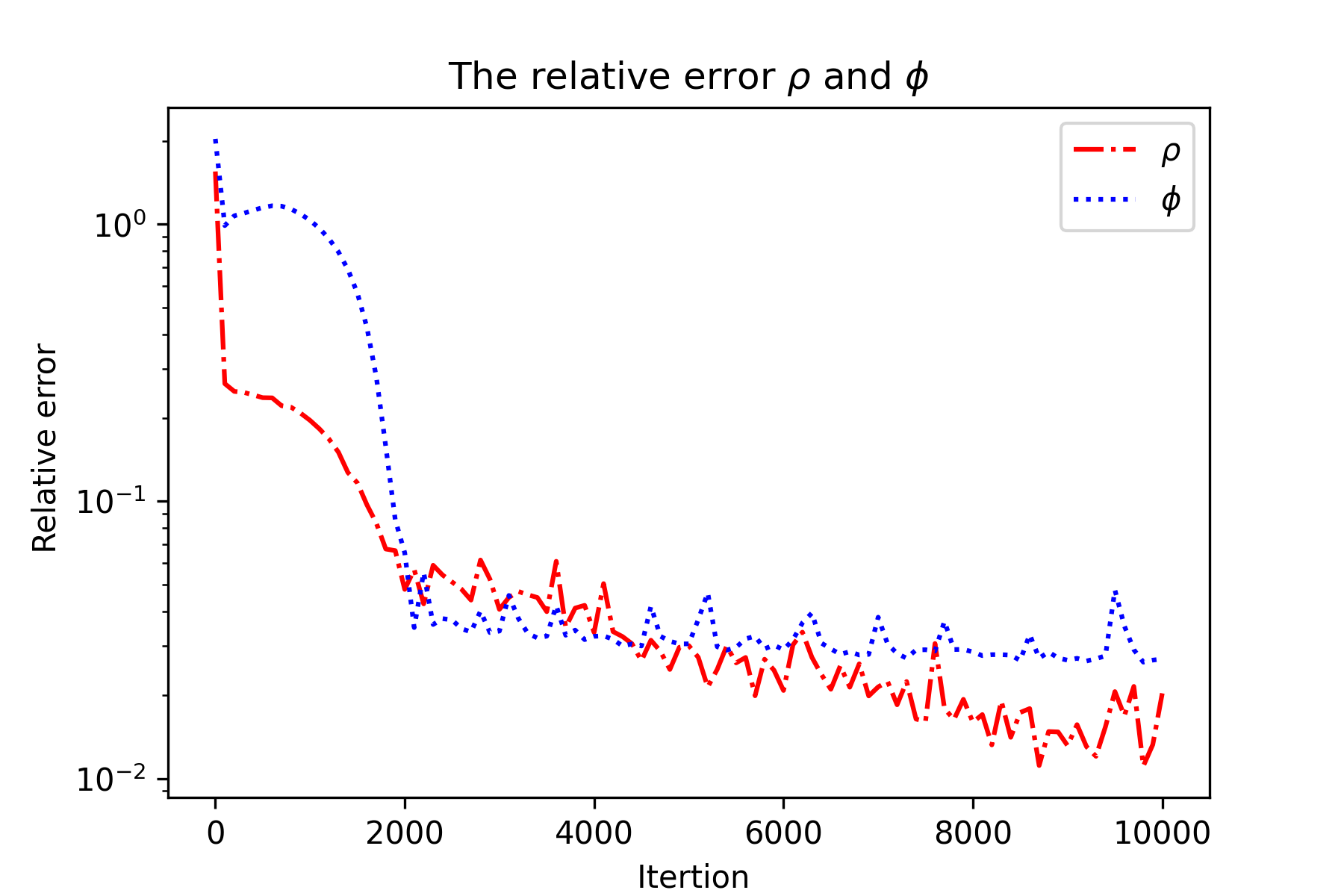}
\centering\includegraphics[width=6.cm]{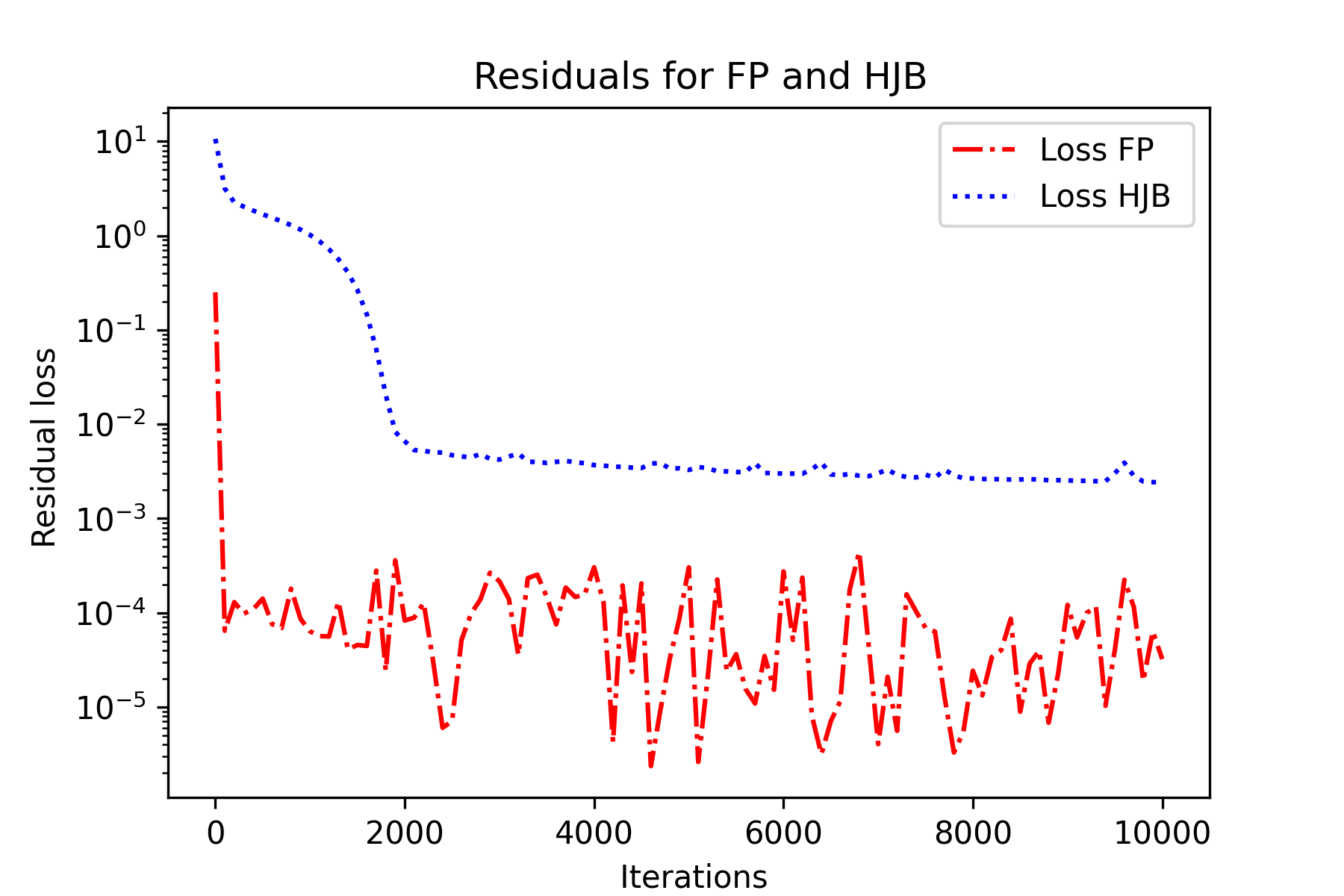}
\caption{The relative error for $\rho$, $\phi$  for the figure on the left. On the right, the HJB, FP Loss.} 
\label{fig2}
\end{figure}

{\bf Test 1:} We consider the system of PDEs [\ref{test}] in one dimension ($d=1$). To obtain results, we run Algorithm [\ref{alg1}] for $10^4$ iterations, using a minibatch of 256 samples at each iteration. The neural networks employed have three hidden layers with 100 neurons each, and utilize the Softplus activation function for $N_{\omega}$ and the Tanh activation function for $N_{\theta}$. Both networks use ADAM with a learning rate of $10^{-4}$, and a weight decay of $10^{-3}$. We employ ResNet as the architecture of the neural networks, with a skip connection weight of 0.5. The numerical results are shown in Figure \ref{fig1}, which compares the approximate solutions obtained by MFDGM to the exact solutions at different time states.\\
To evaluate the performance of MFDGM, we compute the relative error between the model predictions and the exact solutions on a $100 \times 100$ grid within the domain $[0,1]\times[-2,2]$. Additionally, we plot the HJB and FP residual loss, as defined in Algorithm [\ref{alg1}], to monitor the convergence of our method (see Figure \ref{fig2}).

\begin{figure}[htbp]
\centering\includegraphics[width=6.cm]{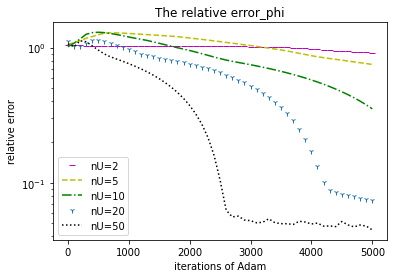}
\centering\includegraphics[width=6.cm]{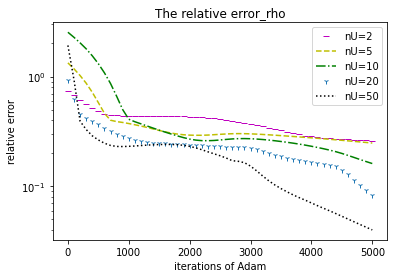}
\caption{The relative error for $\rho$ , $\phi$ in 1-dimension for nU=(2, 5, 10, 20, 50).} 
\label{fig+}
\end{figure}

\begin{figure}[htbp]
\centering\includegraphics[width=6.cm]{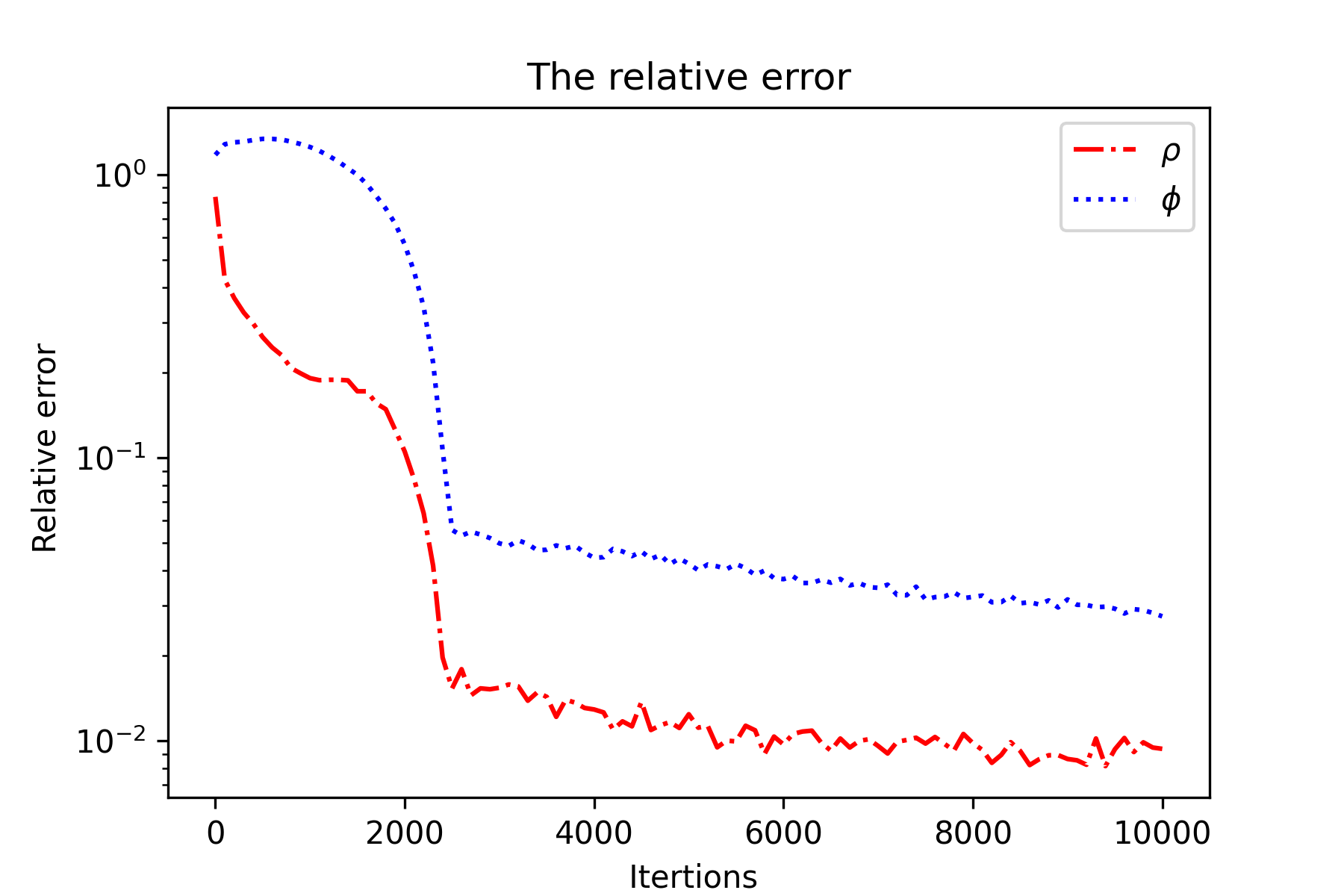}
\centering\includegraphics[width=6.cm]{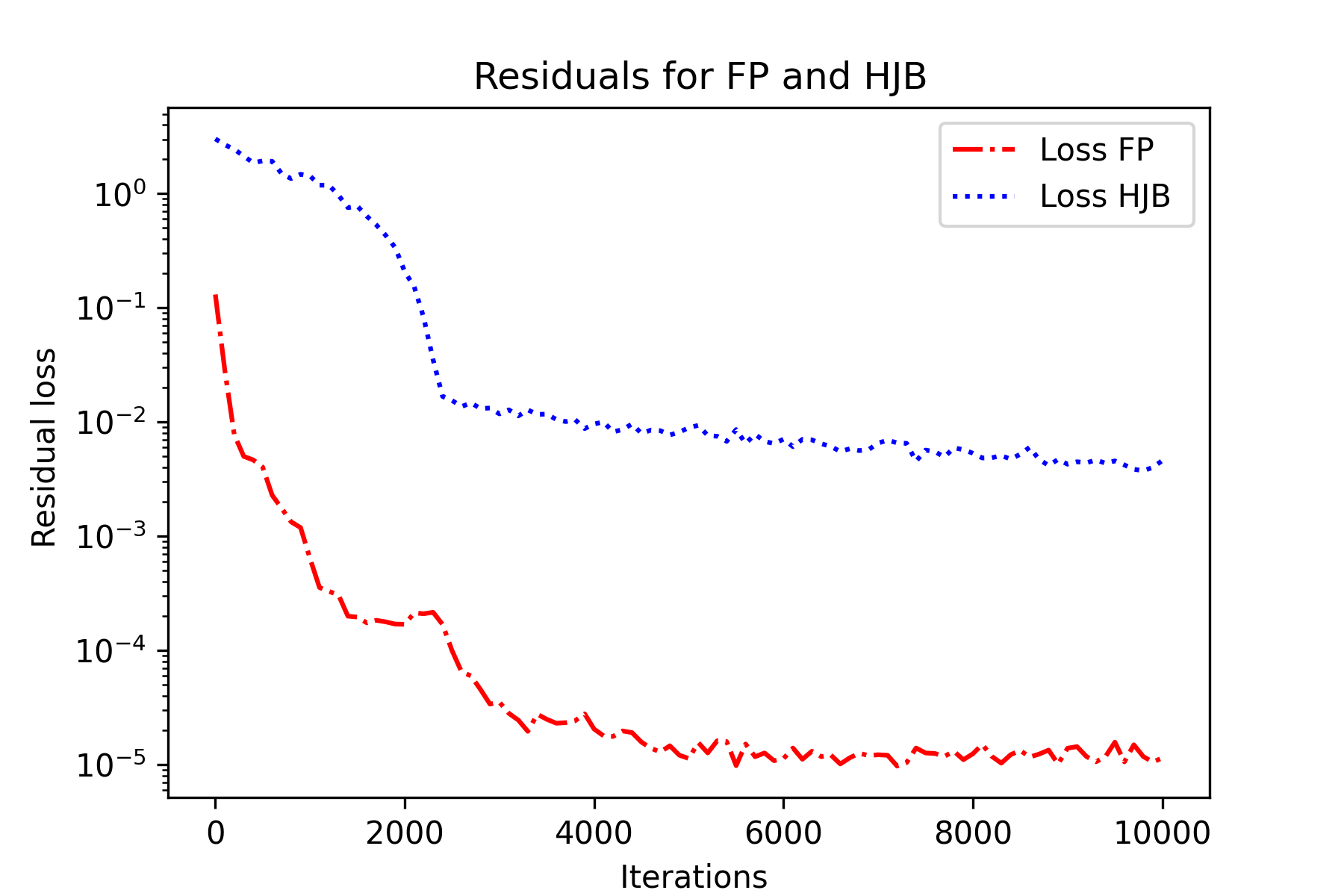}
\caption{Relative error of $\rho$ and $\phi$ and HJB, FP loss for nU=100 after 10000 iterations} 
\label{figr1}
\end{figure}

{\bf Test 2:} In this experiment, we use a single hidden layer with varying numbers of hidden units (nU) for both neural networks. As previously shown in section 2, the number of hidden units can affect the convergence of the model. To verify this, we repeat the previous test using the same hyperparameters and a single hidden layer but with different numbers of hidden units. The relative error between the model predictions and the exact solutions is then calculated on a $100 \times 100$ grid within the domain $[0,1]\times[-2,2]$, as shown in Figure \ref{fig+}. Our objective for this figure was to confirm the result stated in Theorem 1, which suggests that increasing the number of neurons in a single-layer neural network can lead to better results. It is important to note that the performance and robustness of neural networks utilized in the model can be improved by experimenting with diverse training approaches. Selecting the best possible combination of architecture and hyperparameters for the neural networks is essential for achieving the desired outcomes. For example, by increasing the number of neurons and iterations, better results can be obtained, as shown in Figure \ref{figr1}. Therefore, continuous experimentation and optimization of the neural network structure and hyperparameters are necessary to enhance the overall performance and accuracy of the model. Another way to improve the accuracy of the solution is to add appropriate boundary conditions that reflect the physical behavior of the problem. In our work, we have considered only the initial condition and have not explored the effect of boundary conditions on the accuracy of the solution. However, we believe that incorporating appropriate boundary conditions can certainly help to improve the results further. \\
 
\begin{figure}[htbp]
\centering\includegraphics[width=6.cm]{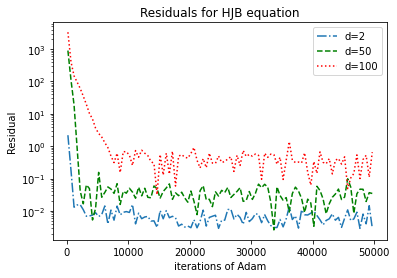}
\centering\includegraphics[width=6.cm]{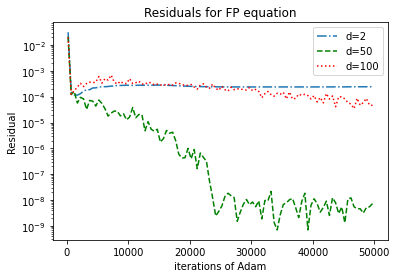}
\caption{The loss HJB and FP equation for d=(2,10,100)} 
\label{fig++}
\end{figure}

\begin{figure}[htbp]
\centering\includegraphics[width=6.cm]{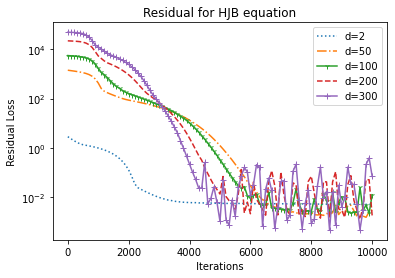}
\centering\includegraphics[width=6.cm]{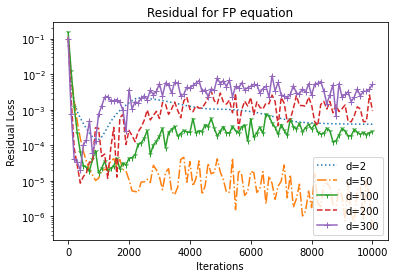}
\caption{The loss HJB and FP equation with a minibatch of 128, 512, and 1024 samples for d=2, d=50, and d=(100,200,300), respectively.} 
\label{fig++++}
\end{figure}

\begin{figure}[htbp]
\centering\includegraphics[width=6.cm]{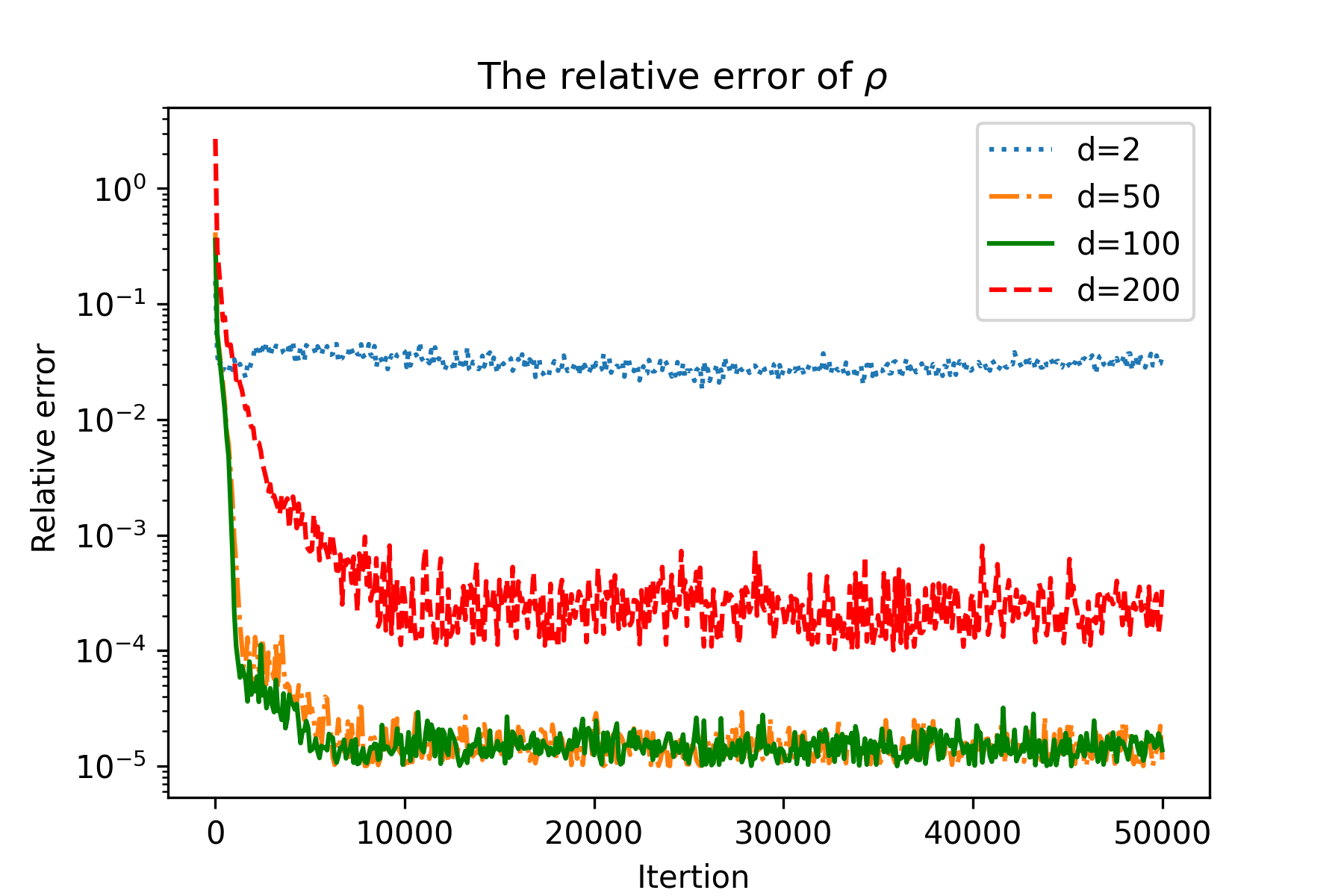}
\centering\includegraphics[width=6.cm]{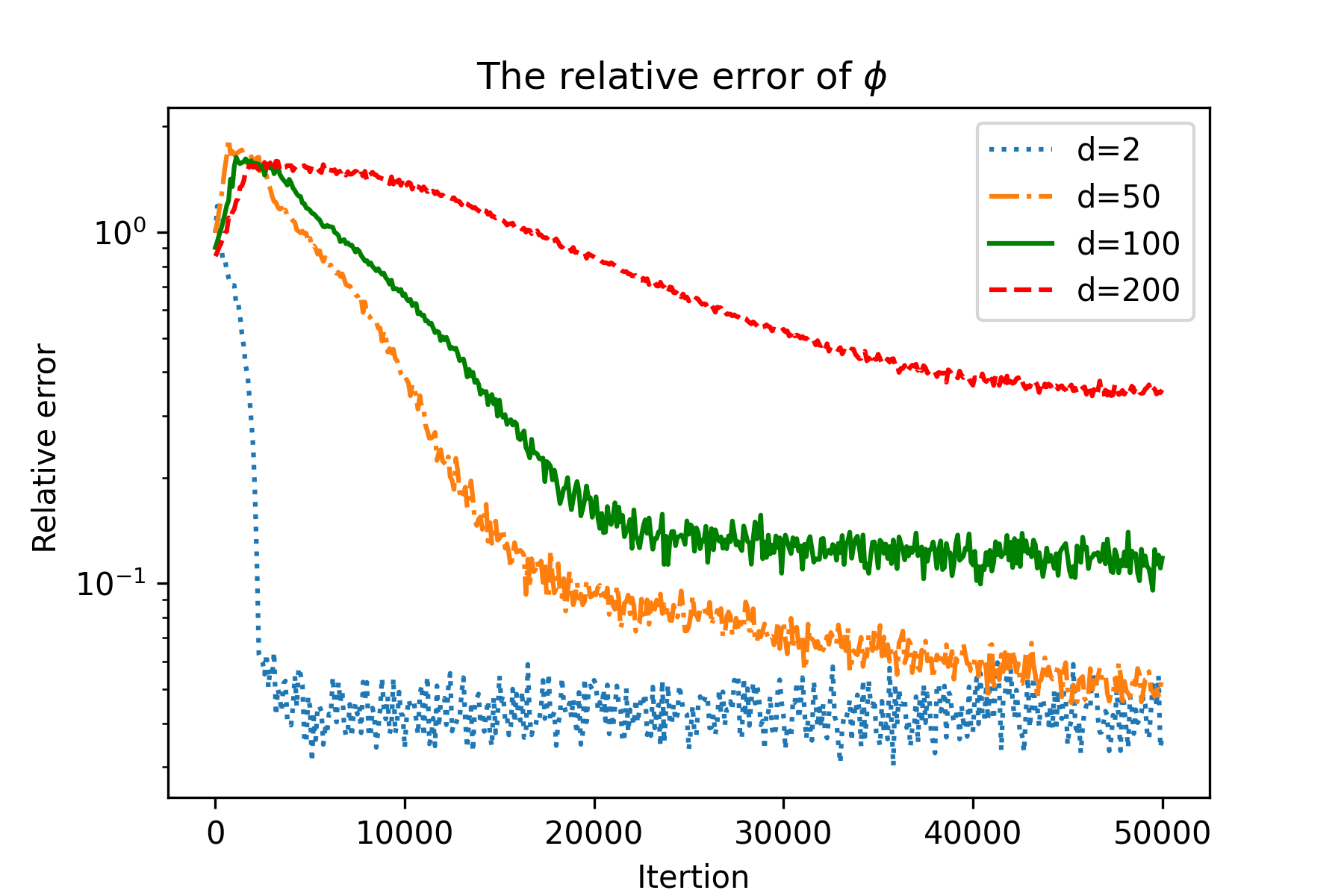}
\caption{The Relative Error of $\rho$ and $\phi$ Using a Minibatch of 1024 Samples for Dimensions 50 and 100.} 
\label{figdim}
\end{figure}

{\bf Test 3:} We conduct two numerical experiments to solve the MFG system (\ref{test}). In the first experiment, we solve the system for dimensions 2, 50, and 100 using neural networks with three hidden layers, each consisting of 100 neurons. Figure \ref{fig++} shows the residuals of the HJB and FP equations over $5.10^4$ iterations, with a minibatch of 1024, 512, and 128 samples used for d=100, d=50, and d=2, respectively. The Softplus activation function is used for $N_{\omega}$ and the Tanh activation function for $N_{\theta}$. Both networks use ADAM with a learning rate of $10^{-4}$, weight decay of $10^{-3}$, and employ ResNet as their architecture with a skip connection weight of 0.5. The results are obtained by recording the residuals every 100 iterations and using a rolling average over 5 points to smooth out the curves.\\
\\
In the second experiment, we increase the dimension to 300 and use a single layer of 100 neurons instead of multiple layers while keeping all other neural network hyperparameters unchanged. This experiment is meant to demonstrate that a single layer can perform better than multiple layers, even when the dimension increases, as seen in section 2. Figure \ref{fig++++} shows improved results compared to the previous experiment, even with fewer iterations, which allows for faster computation times.\\
{\bf Test 4:} We propose another numerical experiment to test the efficiency of our method in high dimensions. We will compute the relative error between the model predictions and the exact solutions over $5.10^4$ iterations Figure \ref{figdim}. Due to hardware limitations, we will only test dimensions  2, 50, 100 and 200 using neural networks with a single hidden layer consisting of 256 neurons. Our aim is to demonstrate the effectiveness of our approach in achieving accurate results in high-dimensional problems, even with limited computational resources. 
\subsection{Comparison }\label{Comparison }
\begin{figure}[htbp]
\centering\includegraphics[width=6.cm]{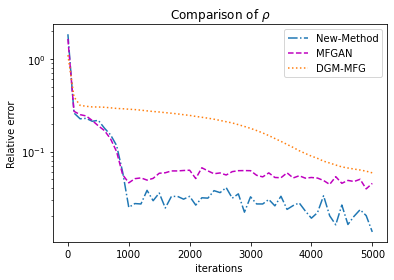}
\centering\includegraphics[width=6.cm]{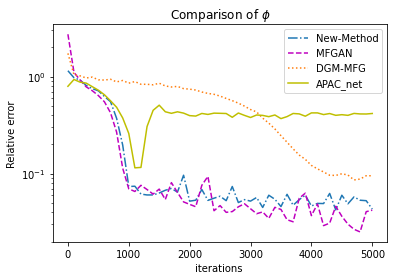}
\caption{comparison between APAC-Net, MFGAN, DGM-MFG, and MFDGM (New-Method).}
\label{fig3}
\end{figure}

In previous sections, we introduced and discussed four methods for solving MFGs: APAC-Net, MFGAN, DGM-MFG, and  MFDGM. Here, we compare these approaches to assess their performance. For APAC-Net, it is only possible to compare the cost values $\phi$ due to the unavailability of the density function. In APAC-Net, the generator neural network represents $\rho$, which generates the distribution. In order to compare the results, we need to use kernel density estimation to transform the distribution into a density, which is only an estimate. We use the simple example from the analytic solution with $d=1$ and $T=1$ for this comparison. The two neural networks in this comparison have three hidden layers with 100 neurons each, and utilize ResNet as their architecture with a skip connection weight of 0.5. They also use the Softplus activation function for $N_{\omega}$ and the Tanh activation function for $N_{\theta}$. For training APAC-Net, MFGAN, and MFDGM, we use ADAM with a learning rate of $10^{-4}$ and a weight decay of $10^{-3}$ for both networks. For training DGM-MFG, we use SGD initialized with a value of $10^{-3}$ and a weight decay of $10^{-3}$ for both networks.\\
We run the four algorithms for $5.10^3$ iterations, using a minibatch of 50 samples at each iteration. The relative error between the model predictions and the exact solutions is then calculated on a $100 \times 100$ grid within the domain $[0,1]\times[-2,2]$, as shown in Figure \ref{fig3}. Our findings indicate that our proposed method exhibits better convergence compared to the
other methods.
\subsection{Application (Traffic Flow):}\label{secLWR} 

\begin{figure}[htbp]    
\centering\includegraphics[width=9.cm]{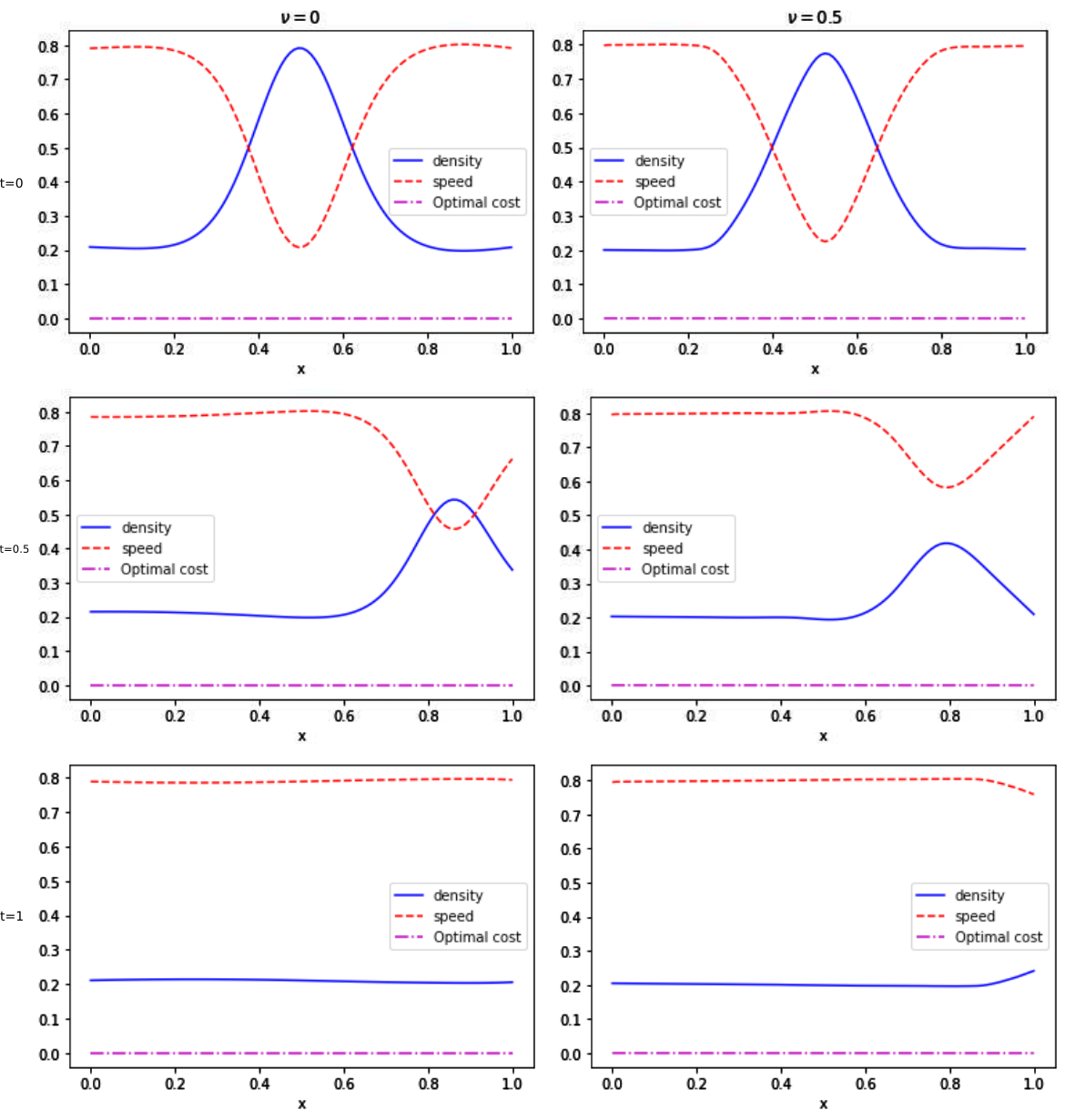}
\caption{The solution of the problem MFG-LWR by MFDGM for ($\nu=0$) and ($\nu=0.5$) at t=(0,0.5,1).} 
\label{fig4}
\end{figure}

In a study published in \cite{huang2019game}, the authors focused on the longitudinal speed control of autonomous vehicles. They developed a mathematical model called a Mean Field Game (MFG) to solve a traffic flow problem for autonomous vehicles and demonstrated that the traditional Lighthill-Whitham-Richards (LWR) model can be used as a solution to the MFG-LWR model described by the following system of equations:
\\
\begin{equation}\label{5l}
MFG-LWR\left\{
\begin{array}{rrrrr}
 V_t+U(\rho)V_x-\frac{1}{2}V_x^2=0,\\
 \rho_t + (\rho u)_{x}=0,\\
 u=U(\rho)-V_x,\\\
V_T=g(\cdot,\rho_T), \ \ \ \ \rho(\cdot,0)=\rho_0.&
\end{array}
\right.
\vspace{-0.25cm}
\end{equation}\\
Here, $\rho$, $V$, and $u$ represent the density, optimal cost, and speed function, respectively, and the Greenshields density-speed relation is given by $U(\rho)=u_{max}(1-\rho/\rho_{jam})$, where $\rho_{jam}$ is the jam density and $u_{max}$ is the maximum speed. By setting $\rho_{jam}=1$ and $u_{max}=1$, the authors generalized the MFG-LWR model to include a viscosity term $\mu>0$, resulting in the following system:
\begin{equation}\label{5}
MFG-LWR\left\{
\begin{array}{rrrrr}
 V_t+\nu \Delta V-H(x,p,\rho)=0,\\
 \rho_t -\nu \Delta \rho-\operatorname{div}(\nabla_pH(x,p,\rho)\rho)=0,\\
V_T=g(\cdot,\rho_T), \ \ \ \ \rho(\cdot,0)=\rho_0.&
\end{array}
\right.
\vspace{-0.25cm}
\end{equation}\\
In this model, $\rho$ and $V$ represent the density and optimal cost function, respectively, and $H$ is the Hamiltonian with a non-separable structure given by 
\begin{equation}\label{nonsep}
H(x,p,\rho)=\frac{1}{2}||p||^2-(1-\rho)p, \ \ \ with \ \ p=V_x,  
\end{equation}
where $p=V_x$. The authors employed a multigrid preconditioned Newton's finite difference algorithm to solve the system in (\ref{5}) for the deterministic case ($\nu = 0$), utilizing a numerical method that involves a finite number of discretization points to reduce computational complexity. It is worth noting that for the stochastic case, traditional methods such as finite difference and finite element methods can be employed to address this problem. Our motivation for adopting a neural network-based approach is not to challenge or replace these traditional methods. Instead, we aim to showcase the capabilities and potential of our approach in efficiently handling the complexities associated with the problems characterized by non-separable Hamiltonians. Therefore, we chose to begin with this well-studied one-dimensional traffic flow problem. This problem serves as an excellent benchmark due to its characteristic of having a non-separable Hamiltonian (\ref{nonsep}), which poses challenges for recent neural network-based methods like Generative Adversarial Networks (GANs). The spatial domain is defined as $\Omega=[0,1]$ and final time $T=1$. The terminal cost $g$ is set to zero and the initial density $\rho_0$ is given by a Gaussian distribution, $\rho_0(x)=0.2-0.6 \ exp\left(\frac{-1}{2}\left(\frac{x-0.5}{0.1}\right)^2\right)$. \\
The corresponding MFG system is,
\begin{equation}\label{1k}
\left\{
\begin{array}{rrrrr}
V_t+\nu \Delta V-\frac{1}{2}||V_x||^2+(1-\rho)V_x=0\\
 \rho_t -\nu \Delta \rho-\operatorname{div}((V_x-(1-\rho))\rho)=0\\
\rho(x,0)=0.2-0.6 \ exp(\frac{-1}{2}(\frac{x-0.5}{0.1})^2), \\ \phi(x,T)=0. 
\end{array}
\right.
\vspace{-0.25cm}
\end{equation}\\
We study the deterministic case $(\nu=0)$ and stochastic case $(\nu=0.5)$. We represent the unknown solutions by two neural networks $N_{\omega}$ and $N_{\theta}$, which have a single hidden layer of 50  neurons. We use the ResNet  architecture with a skip connection weight of 0.5. We employ ADAM with  learning rate $4 \times 10^{-4}$ for $N_\omega$ and  $5 \times 10^{-4}$ for $N_\theta$ and
weight decay of $10^{-4}$ for both networks, batch size 100, in both cases $\nu=0$ and $\nu=0.5$ we use the activation function Softmax and Relu for $N_{\omega}$ and $N_{\theta}$ respectively. In Figure \ref{fig4} we plot over different times the density function, the optimal cost, and  the speed which is calculated according to the density and the optimal cost  \cite{huang2019game} by the following formula,
$$u=u_{max}(1-\rho / \rho_{jam})-V_x$$
where, we take the jam density  $\rho_{jam} =1$ and the maximum speed $u_{max}=1$ and $10^4$ iterations.\\
\begin{figure}[htbp]
\centering\includegraphics[width=6.cm]{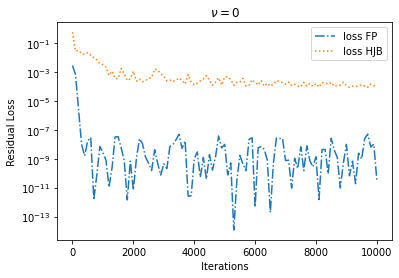}
\centering\includegraphics[width=6.cm]{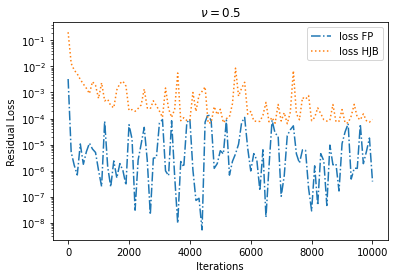}
\caption{The loss HJB and FP equation for ($\nu=0$) and ($\nu=0.5$).}
\label{figloss}
\end{figure}
In Figure (\ref{figloss}),  we plot the HJB, FP residual loss for $\nu=0$ and $\nu=0.5$, which helps
us monitor the convergence of our method. Unfortunately, we do not have the exact solution to compute the error. 
To validate the results of Figure (\ref{fig4}), we use the fundamental traffic flow diagram, an essential tool to comprehend classic traffic flow models. Precisely, this is a graphic that displays a link  between road traffic flux (vehicles/hour) and the traffic density (vehicles/km) \cite{siebel2006fundamental, geroliminis2008existence, keyvan2012exploiting}. We can find this diagram numerically \cite{huang2019game} such as its function $q$ is given by,
$$q(t,x)=\rho(t,x)u(t,x).$$
Figure (\ref{fig5}) shows the fundamental diagram of our results.
\begin{figure}[htbp]
\centering\includegraphics[width=9.cm]{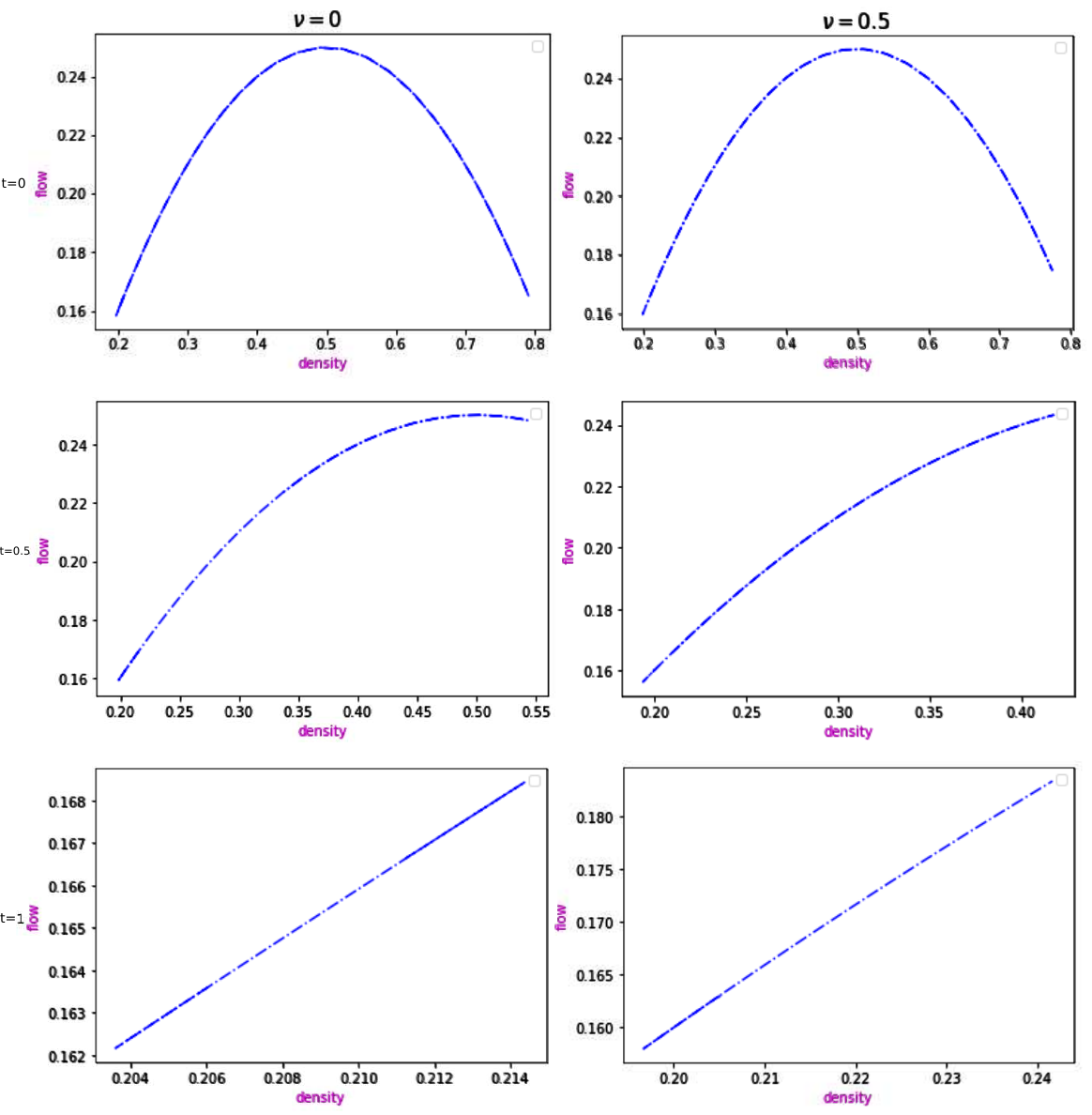}
\caption{Fundamental diagram for $\nu=(0, 0.5)$ at $t=(0, 0.5, 1)$.} 
\label{fig5}
\end{figure}

\section{Conclusion}\label{Conclusion}
\begin{itemize}
    \item We present a new method based on the deep galerkin method (DGM) for solving high-dimensional stochastic mean field games (MFGs). 
    \item We approximate the unknown solutions by two neural networks that were simultaneously trained to satisfy each equation of the MFGs system and forward-backward conditions.
    \item Our method handles up to 300+ dimensions with a single layer for faster computing.
    \item We proved that as the number of hidden units increases, the neural networks converge to the MFG solution.
    \item Comparison with the previous methods shows the efficiency of our approach even with multilayer neural networks. 
    \item Test on traffic flow problem in the deterministic case gives results similar to the newton iteration method, showing that it can solve this problem in the stochastic case.
\end{itemize}
To address the issue of high dimensions in the problem, we used a neural network but found that it took a significant amount of time. While our approach has helped to reduce the time required, it is still not fast enough. Therefore, we are seeking an alternative to neural networks in future research to improve efficiency.
\appendix
\section{Proof of Theorem 3.1.}
\label{Theorem 3.1.}
Denote $\mathcal{N}(\sigma)$ the space of all functions implemented by such a network with a single hidden layer and $n$ hidden units, where $\sigma$ in $\mathcal{C}^{2}\left(\mathbb{R}^{d+1}\right)$ non-constant, and bonded. By {\bf(H1)} we have that  for all $\rho, \phi \in \mathcal{C}^{1,2}\left([0, T] \times \mathbb{R}^{d}\right)$ and $\varepsilon_1, \varepsilon_2>0$, There is $\rho_{\theta}, \phi_{\omega} \in \mathcal{N}(\sigma)$ such That,
\begin{equation}\label{A.1}
    \begin{split}
       &\sup _{(t, x) \in E}\left|\partial_{t} \phi(t, x)-\partial_{t} \phi_{\omega}(t, x)\right|
       \\& \quad\quad+\max _{|a| \leq 2} \sup _{(t, x) \in E}\left|\partial_{x}^{(a)} \phi(t, x)-\partial_{x}^{(a)} \phi_{\omega}(t, x )\right|<\epsilon_1
    \end{split}
\end{equation}
\begin{equation}\label{A.2}
    \begin{split}
       &\sup _{(t, x) \in E}\left|\partial_{t} \rho(t, x)-\partial_{t} \rho_{\theta}(t, x)\right|
       \\& \quad\quad+\max _{|a| \leq 2} \sup _{(t, x) \in E}\left|\partial_{x}^{(a)} \rho(t, x)-\partial_{x}^{(a)} \rho_{\theta}(t, x )\right|<\epsilon_2 
    \end{split}
\end{equation}
From {\bf (H3)} we have that $(\rho, p) \mapsto H( x, \rho, p)$ is locally Lipschitz continuous in $(\rho, p)$, with Lipschitz constant
that can have at most polynomial growth in $\rho$ and $p$, uniformly with respect to $t, x.$ This means that 
\begin{equation*}
    \begin{split}
     |H(x, \rho, p)-H(x, \gamma, s)| \leq&\Big(|\rho|^{q_{1} / 2}+|p|^{q_{2} / 2}+|\gamma|^{q_{3} / 2}+|s|^{q_{4} / 2}\Big)\\
     &\quad\times (|\rho-\gamma|+|p-s|) .   
    \end{split}
\end{equation*}
with some constants $0 \leq q_{1}, q_{2}, q_{3}, q_{4}<\infty$. As a result, we get using Hölder inequality with $\operatorname{exponents} r_{1}, r_{2}$,
$$
\begin{aligned}
& \int_{E}\left|H\left(x, \rho_{\theta}, \nabla_{x} \phi_{\omega}\right)-H\left(x, \rho, \nabla\phi\right)\right|^{2} d \mu_{1}(t, x) \\
&\leq  \int_{E}\left(|\rho_{\theta}(t, x)|^{q_{1}}+\left|\nabla \phi_{\omega}(t, x)\right|^{q_{2}}+|\rho(t, x)|^{q_{3}}+\left|\nabla \phi(t, x)\right|^{q_{4}}\right) \\
&\quad  \times\left(|\rho_{\theta}(t, x)-\rho(t, x)|^{2}+\left|\nabla \phi_{\omega}(t, x)-\nabla \phi(t, x)\right|^{2}\right) d \mu_{1}(t, x) \\
&\leq \Big(\int_{E}(|\rho_{\theta}(t, x)|^{q_{1}}+\left|\nabla \phi_{\omega}(t, x)\right|^{q_{2}}+|\rho(t, x)|^{q_{3}}+\left|\nabla \phi(t, x)\right|^{q_{4}})^{r_{1}} d \mu_{1}(t, x)\Big)^{1 / r_{1}}\\ &\quad\times\Big(\int_{E}(|\rho_{\theta}(t, x)-\rho(t, x)|^{2}+\left|\nabla \phi_{\omega}(t, x)-\nabla \phi(t, x)\right|^{2})^{r_{2}} d \mu_{1}(t, x)\Big)^{1 / r_{2}} \\
&\leq  C_1\Big(\int_{E}(|\rho_{\theta}(t, x)-\rho(t, x)|^{q_{1}}+\left|\nabla \phi_{\omega}(t, x)-\nabla \phi(t, x)\right|^{q_{2}}\\
&\quad\quad+|\rho(t, x)|^{q_{1} \vee q_{3}}+\left|\nabla \phi(t, x)\right|^{q_{2} \vee q_{4}})^{r_{1}} d \mu_{1}(t, x)\Big)^{1 / r_{1}} \\
&\quad \times\Big(\int_{E}(|\rho_{\theta}(t, x)-\rho(t, x)|^{2}+|\nabla \phi_{\omega}(t, x)-\nabla \phi(t, x)|^{2})^{r_{2}} d \mu_{1}(t, x)\Big)^{1 / r_{2}} \\
&\leq C_1\left(\epsilon_1^{q_{1}}+\epsilon_2^{q_{2}}+\sup _{E}|\rho|^{q_{1} \vee q_{3}}+\sup _{E}\left|\nabla \phi\right|^{q_{2} \vee q_{4}}\right) (\epsilon_1^2+\epsilon_2^2)\\
&\leq  C_1(\epsilon_1^2+\epsilon_2^2) \leq  C_1\epsilon^2,
\end{aligned}
$$
where the constant $C_1 < \infty$ may change from line to line and  $q_i \vee q_j = max\{q_i, q_j\}$. In the two last steps we used \ref{A.1}, \ref{A.2} and {\bf (H2)}.
We recall that,
\begin{equation*}
\mathcal{H}_1(\rho_{\theta},\phi_{\omega})=\partial_t \phi_{\omega}(t,x)
     + \nu \Delta \phi_{\omega}(t,x)- H(x,\rho_{\theta}(t,x),\nabla \phi_{\omega}(t,x)).
\end{equation*}
Note that $\mathcal{H}_1(\rho,\phi)=0$ for $\rho, \theta$ that solves the system of PDEs,
$$
\begin{aligned}
L_1(\rho_{\theta},\phi_{\omega})=&\Big\|\mathcal{H}_1(\rho_{\theta},\phi_{\omega})\Big\|^2_{L^2(E)}+ \Big\|\phi_{\omega}(T,x)-\phi(T,x)\Big\|^2_{L^2(\Omega)} \\
=&\Big\|\mathcal{H}_1(\rho_{\theta},\phi_{\omega}) -\mathcal{H}_1(\rho,\phi)\Big\|^2_{L^2(E)}+\Big\|\phi_{\omega}(x,T)-g(x,\rho_{\theta}(x,T))\Big\|^2_{L^2(\Omega)}  \\
\leq & \int_{E}\left|\partial_{t} \phi_{\omega}(t, x)-\partial_{t} \phi(t, x)\right|^{2} d \mu_{1}(t, x)\\ &+|\nu|\int_{E}\left|\Delta\phi_{\omega}(t,x)-\Delta\phi(t,x)  \right|^2 d \mu_{1}(t, x)\\
&+\int_{E}\left|H\left(x, \rho_{\theta}, \nabla \phi_{\omega}\right)-H\left(x, \rho, \nabla \phi\right)\right|^{2} d \mu_{1}(t, x) \\
&+\int_{\Omega}|\phi_{\omega}(T,x)-\phi(T, x)|^{2} d \mu_{2}(t, x) \\
\leq & C_1 (\epsilon_1^{2}+\epsilon_2^{2})\leq  C_1\epsilon^2
\end{aligned}
$$
for an appropriate constant $C_1<\infty$. In the last step, we use \ref{A.1}, \ref{A.2} and the previous result.\\
For $L_2$ we use remark \ref{rmk} to simplified the nonlinear term,
\begin{equation*}
    \begin{split}
    \operatorname{div}(\rho \nabla_pH(x,\rho,\nabla\phi))&=\alpha_1(x,\rho,\nabla\phi) + \alpha_2(x,\rho,\nabla\phi)+\alpha_3(x,\rho,\nabla\phi),
    \end{split}
\end{equation*}    
where,
\begin{equation*} 
    \begin{split} \alpha_1(x,\rho,\nabla\phi)&=\nabla_pH(x,\rho,\nabla\phi)\nabla\rho,\\
\alpha_2(x,\rho,\nabla\phi)&=\nabla_{p\rho}H(x,\rho,\nabla\phi)\nabla\rho.\rho,\\
\alpha_3(x,\rho,\nabla\phi)&=\sum_{i,j}\nabla_{p_ip_j}H(x,\rho,\nabla\phi)(\partial_{x_jx_i}\phi)\rho. \end{split}
\end{equation*}
In addition, from {\bf (H3)}  we have also $\nabla_pH( x, \rho, p),$ $\nabla_{p\rho}H( x, \rho, p),$ and  $\nabla_{pp}H( x, \rho, p)$ are locally Lipschitz continuous in $(\rho, p)$. Then, we have after an application of Holder inequality, for some constant $C_2< \infty$ that may change from line to line,
\begin{equation*}
    \begin{split}
      &\int_{E}\left|\alpha_{1}\left(x, \rho_{\theta}, \nabla \phi_{\omega}\right)-\alpha_{1}(x, \rho, \nabla \phi)\right|^{2} d \mu_{1}(t, x)\\
&=\int_{E}\left|\nabla_{p_\omega} H\left(x, \rho_{\theta}, \nabla \phi_{\omega}\right)\nabla \rho_{\theta}-\nabla_{p} H(x, \rho, \nabla \phi)\nabla \rho\right|^{2} d \mu_{1}(t, x)\\
&\leq \int_{E} \Big | \Big(\nabla_{p_{\omega}} H\left(x, \rho_{\theta}, \nabla \phi_{\omega}\right)-\nabla_{p} H(x, \rho, \nabla \phi)\Big) \nabla\rho \Big|^{2} d \mu_{1}(t, x)\\
& \quad \quad +\int_{E}\Big|\nabla_{p_{\omega}} H\left(x, \rho_{\theta}, \nabla \phi_{\omega}\right)\left(\nabla\rho_{\theta}-\nabla \rho\right)\Big|^{2} d \mu_{1}(t, x)\\
&\leq C_{2}\left(\int_{E}\Big|\nabla_{p_{\omega}} H\left(x, \rho_{\theta}, \nabla \phi_{\omega}\right)-\nabla_{p} H(x, \rho, \nabla \phi)\Big|^{2 r_1} d \mu_{1}\left(t,x \right)\right)^{1 / r_1}\\
& \quad \quad  \times \Big(\int_{E}|\nabla\rho|^{2 r_2} d \mu_{1}(t, x)\Big)^{1 / r_2}+C_2 \left(\int_{E}\Big|\nabla_{p_{\omega}} H\left(x, \rho_{\theta}, \phi_{\omega}\right)\Big|^{2 s_1} d \mu_{1}(t, x)\right)^{1 / s_1}\\ 
& \quad \quad\quad\quad\times\left(\int_{E}\left|\nabla\rho_{\theta}-\nabla\rho\right|^{2 s_2} d \mu_{1}(t, x)\right)^{1 / s_2}\\
&\leq  C_2\Big(\int_{E}|\nabla\rho|^{2 r_2} d \mu_{1}(t, x)\Big)^{1 / r_2}\\ &\quad \quad\times\Big(\int_{E}(|\rho_{\theta}(t, x)-\rho(t, x)|^{q_{1}}+\left|\nabla \phi_{\omega}(t, x)-\nabla \phi(t, x)\right|^{q_{2}}\\
& \quad \quad\quad +|\rho(t, x)|^{q_{1} \vee q_{3}}+\left|\nabla \phi(t, x)\right|^{q_{2} \vee q_{4}})^{v_1r_{1}} d \mu_{1}(t, x)\Big)^{1 / v_1r_{1}} \\
& \quad  \quad \times\Big(\int_{E}(|\rho_{\theta}(t, x)-\rho(t, x)|^{2}+\left|\nabla_{x} \phi_{\omega}(t, x)-\nabla_{x} \phi(t, x)\right|^{2})^{v_2r_{2}} d \mu_{1}(t, x)\Big)^{1 / v_2r_{2}} \\
&\quad\quad+C_2 \left(\int_{E}\Big|\nabla_{p_{\omega}} H\left(x, \rho_{\theta}, \phi_{\omega}\right)\Big|^{2 s_1} d \mu_{1}(t, x)\right)^{1 / s_1}\times\left(\int_{E}\left|\nabla\rho_{\theta}-\nabla\rho\right|^{2 s_2} d \mu_{1}(t, x)\right)^{1 / s_2}\\
&\leq C_2 (\epsilon_1^2+\epsilon_2^2)  \leq  C_1\epsilon^2
\end{split}
\end{equation*}
where in the last steps, we followed the computations previously. We do same for $\alpha_2(x,\rho,\nabla\phi)$ and $\alpha_3(x,\rho,\nabla\phi),$ we obtain for a $C_2<\infty$,
\begin{equation*}
    \begin{split}
        &\int_{E}\Big|\operatorname{div}(\rho_{\theta} \nabla_{p_{\omega}}H(x,\rho_{\theta},\nabla\phi_{\omega}))- \operatorname{div}(\rho \nabla_pH(x,\rho,\nabla\phi))\Big|^{2} d \mu_{1}(t, x)\\
        &\leq C_2 (\epsilon_1^2+\epsilon_2^2) \leq  C_1\epsilon^2.
    \end{split}
\end{equation*}
We recall that,
\begin{equation*}
\mathcal{H}_2(\rho_{\theta},\phi_{\omega})=\partial_t \rho_{\theta}(t,x)
     - \nu \Delta \rho_{\theta}(t,x)- \operatorname{div} \left(\rho_{\theta}(t,x)\nabla_pH(x,\rho_{\theta}(t,x),\nabla \phi_{\omega}(t,x))\right)
\end{equation*}
Note that $\mathcal{H}_2(\rho,\phi)=0$ for $\rho, \theta$ that solves the system of PDEs, then we have,
$$
\begin{aligned}
L_2(\rho_{\theta},\phi_{\omega})
=&\Big\|\mathcal{H}_2(\rho_{\theta},\phi_{\omega})\Big\|^2_{L^2(E)}
+ \Big\|\rho_{\theta}(0,x)-\rho_0(x)\Big\|^2_{L^2(\Omega)}\\
=&\Big\|\mathcal{H}_2(\rho_{\theta},\phi_{\omega}) -\mathcal{H}_2(\rho,\phi)\Big\|^2_{L^2(E)}+\Big\|\rho_{\theta}(0,x)-\rho_0(x)\Big\|^2_{L^2(\Omega)}\\
\leq & \int_{E}\left|\partial_{t} \rho_{\theta}(t, x)-\partial_{t} \rho(t, x)\right|^{2} d \mu_{1}(t, x)\\ &+|\nu|\int_{E}\left|\Delta\rho_{\theta}(t,x)-\Delta\rho(t,x)  \right|^2 d \mu_{1}(t, x)\\
&+\int_{E}\Big|\operatorname{div}(\rho_{\theta} \nabla_{p_{\omega}}H(x,\rho_{\theta},\nabla\phi_{\omega}))- \operatorname{div}(\rho \nabla_pH(x,\rho,\nabla\phi))\Big|^{2} d \mu_{1}(t, x)\\
&+\int_{\Omega}|\rho_{\theta}(0,x)-\rho_0(x)|^{2} d \mu_{2}(t, x) \\
\leq & C_2 (\epsilon_1^{2}+\epsilon_2^{2})\leq  C_1\epsilon^2,
\end{aligned}
$$
for an appropriate constant $C_2<\infty$. The proof
of theorem \ref{the} is complete after rescaling $\epsilon$ \\
\section{Proof of Theorem 3.2.}
\label{Theorem 3.2.}
We follow the method used in \cite{sirignano2018dgm} for a single PDE. (See also section 4 in \cite{gallego2007existence} for a coupled system). Let us denote the solution of  problem \ref{12} by. $\left(\hat{\rho}_{\theta}^{n}, \hat{\phi}_{\omega}^{n}\right) \in V=V_0^{2,2} \times V_0^{2,2}$. Due to Conditions $\left(H_{4}\right)-\left(H_{6}\right)$ and by using  lemma 1.4 \cite{porzio1999existence} on each equation then, there exist, $C_{1}$, $C_{2}$ such that:
$$\|\hat{\rho}_{\theta}^{n}\|_{V_0^{2,2}} \leq C_1$$
$$\|\hat{\phi}_{\omega}^{n}\|_{V_0^{2,2}} \leq C_2$$
These applies and gives that the both sequence $\{\hat{\rho}_{\theta}^n\}_{n\in\mathbf{N}}$, $\{ \hat{\phi}_{\omega}^n\}_{n\in\mathbf{N}}$ are uniformly bounded
with respect to n in at least $V$. These uniform
energy bounds imply the existence of two subsequences, (still denoted in
the same way) $\{\hat{\rho}_{\theta}^n\}_{n\in\mathbf{N}}$, $\{ \hat{\phi}_{\omega}^n\}_{n\in\mathbf{N}}$ and two functions $\rho$, $\phi$ in $L^{2}\left(0, T ; W_{0}^{1,2}(\Omega)\right)$ such that,
\begin{equation*}
\hat{\rho}_{\theta}^n \rightarrow \rho \text { weakly in } L^{2}\left(0, T: W_{0}^{1, 2}(\Omega)\right)
\end{equation*}
\begin{equation*}
\hat{\phi}_{\omega}^n \rightarrow \phi \text { weakly in } L^{2}\left(0, T: W_{0}^{1, 2}(\Omega)\right)
\end{equation*}
Next let us set $q=1+\frac{d}{d+4} \in(1,2)$ and note that for conjugates, $r_{1}, r_{2}>1$ such that $1 / r_{1}+1 / r_{2}=1$
$$
\begin{aligned}
\int_{\Omega_T}\left|\gamma\left(t, x, \hat{\rho}_{\theta}^n, \nabla \hat{\phi}_{\omega}^n\right)\right|^{q} & \leq \int_{\Omega_T}|\lambda|^{q}\left|\nabla\hat{\phi}_{\omega}^n\right|^{q}\\
& \leq\left(\int_{\Omega_T}|\lambda|^{r_{1} q}\right)^{1 / r_{1}}\left(\int_{\Omega_T}\left|\nabla \hat{\phi}_{\omega}^n\right|^{r_{2} q} \right)^{1 / r_{2}}
\end{aligned}
$$
Let us choose $r_{2}=2 / q>1$. Then we calculate $r_{1}=\frac{r_{2}}{r_{2}-1}=\frac{2}{2-q}$. Hence, we have that $r_{1} q=d+2$. Recalling the assumption $\lambda \in L^{d+2}\left(E\right)$ and the uniform bound on the $\nabla \hat{\phi}_{\omega}^n$ we subsequently obtain that for $q=1+\frac{d}{d+4}$, there is a constant $C<\infty$ such that
$$
\int_{\Omega_T}\left|\gamma\left(t, x, \hat{\rho}_{\theta}^n, \nabla \hat{\phi}_{\omega}^n\right)\right|^{q}  \leq C
$$
On the other hand, it is obvious that $a_1$  is bounded uniformly then, according to the HJB equation
of \ref{12}, we have $\left\{\partial_{t} \hat{\phi}_{\omega}^n\right\}_{n \in \mathbb{N}}$ is bounded uniformly with respect to $n$  in $L^{2}\left(0, T ; W^{-1,2}(\Omega)\right)$. Then we can extract a subsequence, (still denoted in
the same way) $\left\{\partial_{t} \hat{\phi}_{\omega}^n\right\}_{n \in \mathbb{N}}$ such that
\begin{equation*}
\partial_{t} \hat{\phi}_{\theta}^n \rightarrow \partial_{t} \phi \text { weakly in } L^{2}\left(0, T ; W^{-1,2}(\Omega)\right)
\end{equation*}
Also, it will be shown that
\begin{equation*}
\partial_{t} \hat{\rho}_{\theta}^n \rightarrow \partial_{t} \rho \text { weakly in } L^{2}\left(0, T ; W^{-1,2}(\Omega)\right)
\end{equation*}
Since the problem is nonlinear, the weak convergence of $\hat{\phi}_{\omega}^n$ and $\hat{\rho}_{\theta}^n$ in the space $L^{2}\left(0, T ; W_{0}^{1, 2}(\Omega)\right)$ is not enough in order to prove that $\phi$ and $\rho$ are a solution of problem \ref{11}. To do this, we need the almost everywhere convergence of the gradients for a subsequence of the approximating solutions $\hat{\phi}_{\omega}^n$ and $\hat{\rho}_{\theta}^n$.\\
However, the uniform boundedness of $\{ \hat{\phi}_{\omega}^n\}_{n\in\mathbf{N}}$ and $\{\hat{\rho}_{\theta}^n\}_{n\in\mathbf{N}}$ in $L^{2}\left(0, T; W_{0}^{1, 2}(\Omega)\right)$  and their weak convergence to $\phi$ and $\rho$ respectively in that space, allows us to conclude, by using Theorem $3.3$ of \cite{boccardo1997nonlinear} on each equation, that
$$\nabla \hat{\phi}_{\omega}^n \rightarrow \nabla \phi \ \ \text{almost \  everywhere \ in} \ \ \Omega_T .$$
$$\nabla \hat{\rho}_{\theta}^n \rightarrow \nabla \rho \ \ \text{almost \  everywhere \ in} \ \ \Omega_T .$$
Hence, we obtain that $\{ \hat{\phi}_{\omega}^n\}_{n\in\mathbf{N}}$ and $\{\hat{\rho}_{\theta}^n\}_{n\in\mathbf{N}}$ converges respectively to $\phi$ and $\rho$ strongly in $L^{p}\left(0, T ; W_{0}^{1, p}(\Omega)\right)$ for every $p<2$.
It remains to discuss the convergence of $\phi_{\omega}^n - \hat{\phi}_{\omega}^n$ and $\rho_{\theta}^n - \hat{\rho}_{\theta}^n$ to zero. By  last step of proof theorem 7.3  \cite{sirignano2018dgm} we get $\left\{\phi_{\omega}^n - \hat{\phi}_{\omega}^n\right\}_{n \in \mathbb{N}}$ and $\left\{\rho_{\theta}^n - \hat{\rho}_{\theta}^n\right\}_{n \in \mathbb{N}}$ goes to zero strongly in $L^{p}\left(\Omega_T\right)$ for every $p<2$. Finally we conclude the proof of the convergence in $L^p\left(\Omega_T\right)$ for every $ p < 2$

 \bibliographystyle{elsarticle-num} 
 \bibliography{elsarticle-template-num}





\end{document}